\documentclass{article}

\usepackage{microtype}
\usepackage{graphicx}
\usepackage{subfigure}
\usepackage{booktabs} 
\usepackage{amsmath}
\usepackage{multirow}
\usepackage{mathtools}
\usepackage{enumitem}
\usepackage{tablefootnote}
\usepackage{hyperref}

\newtheorem{theorem}{Theorem}

\usepackage[accepted]{icml2021}

\icmltitlerunning{Gibbs with Gradients}
\DeclareUnicodeCharacter{2212}{-}
\begin{document}

\twocolumn[
\icmltitle{Oops I Took A Gradient: Scalable Sampling for Discrete Distributions}



\icmlsetsymbol{equal}{*}

\begin{icmlauthorlist}
\icmlauthor{Will Grathwohl}{to,goo}
\icmlauthor{Kevin Swersky}{goo}
\icmlauthor{Milad Hashemi}{goo}
\icmlauthor{David Duvenaud}{to,goo}
\icmlauthor{Chris J. Maddison}{to}
\end{icmlauthorlist}

\icmlaffiliation{to}{University of Toronto and Vector Institute}
\icmlaffiliation{goo}{Google Research, Brain Team}

\icmlcorrespondingauthor{Will Grathwohl}{wgrathwohl@cs.toronto.edu}

\icmlkeywords{Machine Learning, ICML, MCMC, Energy-Based Models, Sampling, Discrete}

\vskip 0.3in
]



\printAffiliationsAndNotice{}  

\begin{abstract}
We propose a general and scalable approximate sampling strategy for probabilistic models with discrete variables.
Our approach uses gradients of the likelihood function with respect to its discrete inputs to propose updates in a Metropolis-Hastings sampler.
We show empirically that this approach outperforms generic samplers in a number of difficult settings including Ising models, Potts models, restricted Boltzmann machines, and factorial hidden Markov models.
We also demonstrate the use of our improved sampler for training deep energy-based models (EBM) on high dimensional discrete data.
This approach outperforms variational auto-encoders and existing energy-based models.
Finally, we give bounds showing that our approach is near-optimal in the class of samplers which propose local updates. 
\end{abstract}

\section{Introduction}
Discrete structure is everywhere in the real world, from text to genome sequences. The scientific community is building increasingly complex models for this discrete data, increasing the need for methods to sample from discrete distributions. Sampling from a discrete distribution may seem like a simpler task than sampling from a continuous one: even a one-dimensional continuous distribution can have an uncountable infinity of outcomes, whereas a discrete distribution is at most countable. However, most common continuous distributions have some kind of simplifying structure, such as differentiable densities, which can be exploited to speed up sampling and inference. 

Of course, many discrete distributions have structure as well. Notably, discrete distributions over combinatorial spaces often have some kind of block independence structure among their variables. Although this can be used to speed up sampling and inference, it may be difficult to detect such structure automatically. Typically, users need to know this structure a priori and must hard-code it into an algorithm to speed up sampling.

\begin{figure}[t]
    \centering
    \includegraphics[width=.48\textwidth]{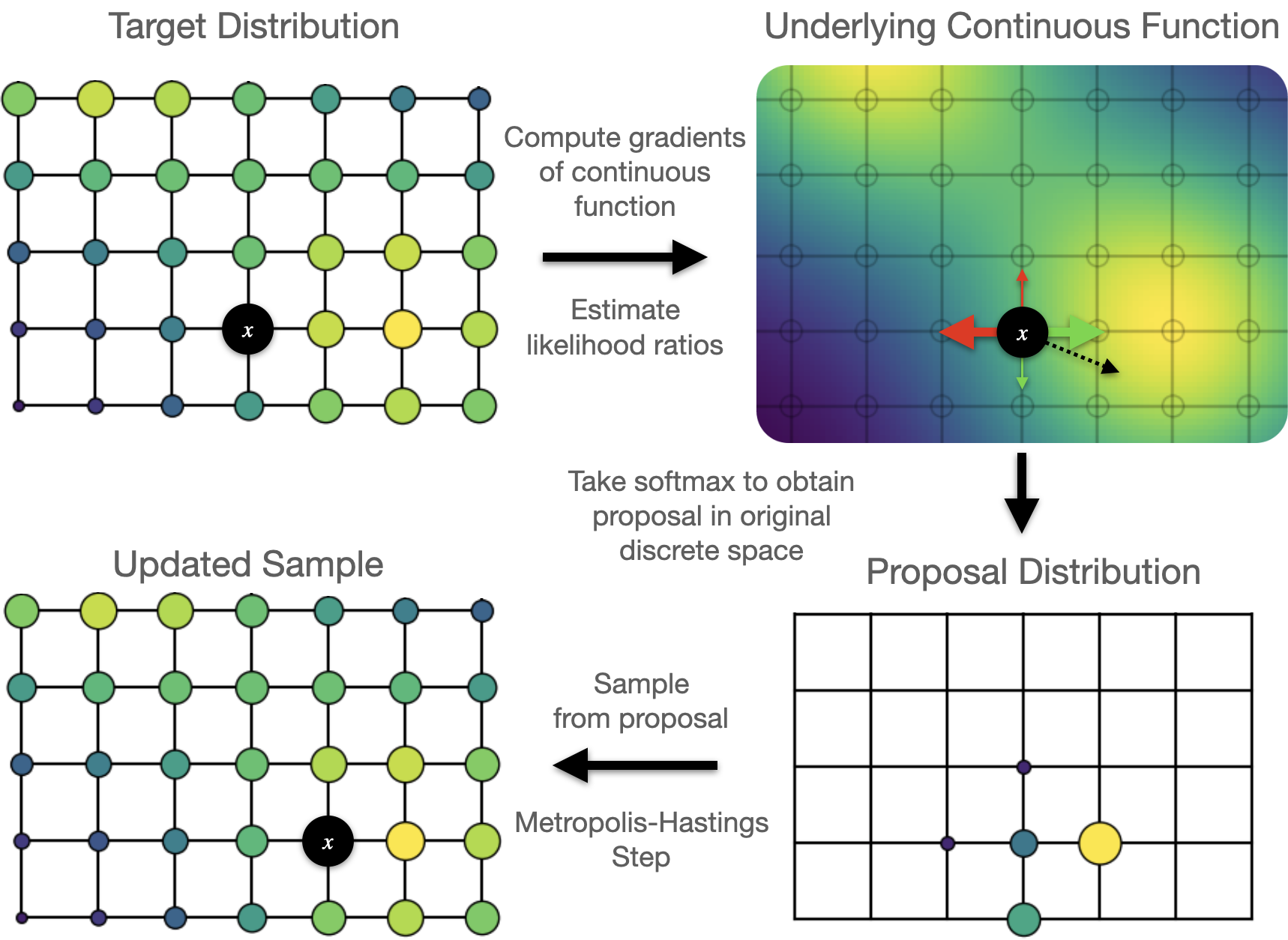}
    \caption{Our approach visualized. Often discrete distributions are defined by continuous functions whose input is restricted to a discrete set; here $R^2$ restricted to $\mathcal{Z}^2$. We use a Taylor series computed on the underlying continuous function to estimate likelihood ratios of making discrete moves; here $\pm 1$ in either direction. These estimated likelihood ratios are used to inform a proposal distribution over moves in the original discrete space.}
    \label{fig:fig1}
\end{figure}

In search for a unifying structure, we notice that many discrete distributions are written (and implemented) as differentiable functions of real-valued inputs. The discrete structure is created by restricting the continuous inputs to a discrete subset of their domain.
In this paper, we use the gradients of these underlying continuous functions to inform proposal distributions for MCMC sampling in large discrete probabilistic models.
This new family of gradient-informed proposals for MCMC may be seen as a class of adaptive Gibbs samplers or a fast approximation to locally-informed proposals \citep{umrigar1993accmet, liu1996peskun, zanella2020informed}.

As we show, this gradient information is cheaply available for many discrete distributions and can lead to orders of magnitude improvements in sampling efficiency. In some cases, it even outperforms samplers that exploit hard-coded independence structure.


We apply these sampling techniques to improve parameter inference in discrete energy-based models such as Ising and Potts models.
We also find that our method is particularly well-suited to sample from unnormalized discrete distributions parameterized by deep neural networks. The improved efficiency of our sampler allows us to apply many techniques from large-scale continuous EBMs to successfully train deep EBMs for discrete data. These models are simple to define, flexible, and outperform baseline variational auto-encoders~\citep{kingma2013auto} and existing energy-based models.


\section{Background}

In this work we are concerned with sampling from unnormalized distributions over discrete data
\begin{align*}
    \log p(x) = f(x) - \log Z
\end{align*}
where $f(x)$ is the unnormalized log-probability of $x$ and ${Z = \sum_{x} e^{f(x)}}$ is the normalizing constant which we assume is unknown. We restrict our study to $D$-dimensional binary $x \in \{0, 1\}^D$ and categorical $x \in \{0, 1, \ldots, K\}^D$ data as all finite-dimensional discrete distributions can be embedded in this way. Throughout the paper we assume all categorical variables are one-hot encoded.

\subsection{Gibbs Sampling}

Gibbs sampling is perhaps the simplest and most generic method for sampling from discrete distributions. At each step, the sampler partitions the dimensions of an input $x$ into two groups $x_u$ and $x_{-u}$ where $u$ is a subset of the $D$ dimensions of $x$ and $-u$ is its complement. The next sample in the chain is created by updating $x_u$ to be a sample from $p(x_u | x_{-u})$, the conditional distribution of the chosen dimensions, given all others. In some distributions, block-independence structure exists, making certain partitions easy to sample from and update in parallel. 

In the worst case, if such structure does not exist, we can let $x_u = x_i$, simply the $i$-th dimension of $x$. In this setting $p(x_i|x_{-i})$ is simply a one-dimensional categorical distribution over $K$ possible outcomes, which requires $K$ evaluations of the unnormalized log-density function. We typically fix an ordering of the dimensions and iterate through this ordering to ensure that each dimension is updated. 


While simple, updating one dimension at a time can be problematic for high-dimensional data. Consider for example a binary model of the MNIST dataset. Almost all dimensions will represent the background and if chosen for a Gibbs update, they will almost certainly not change. Similarly, the pixels on the interior of a digit will also not change. This amounts to wasted computation every time we propose a dimension, which does not change. Ideally, if we could bias our partition choice to dimensions which are likely to change, we could build a more efficient Gibbs sampler. 

Consider a similar sampler for the binary case; Metropolis-Hastings with the proposal distribution ${q(x'|x) = \sum_i q(x'|x, i)q(i)}$ where $q(i)$ is a distribution over indices $i \in \{1, \ldots, D\}$ and $q(x'|x, i) = \delta(x' = x_{-i})$, where $\delta(\cdot)$ is the Dirac-delta distribution. With this sampler we will first sample an index $i \sim q(i)$, then flip the $i$-th bit of $x$ to obtain $x'$ and accept this proposed update with probability 
\begin{align}
    \min\left\{\exp(f(x') - f(x))\frac{q(x|x')}{q(x'|x)} , 1\right\}.
    \label{eq:mh}
\end{align}
This approach can lead to considerable performance improvements if $q(i)$ is biased towards dimensions which are more likely to flip. 
To this end, considerable work has been done to use prior sampling data to adaptively update the $q(i)$ while maintaining the validity of MCMC sampling~\citep{latuszynski2013adaptive, richardson2010bayesian}.


Revisiting our MNIST example, we can see the pixels most likely to change are those at the edge of a digit. Where this edge is varies considerably and depends on the entire input $x$. Thus, we can imagine an input dependent proposal $q(i|x)$ should be able to outperform an unconditional proposal $q(i)$.
Our proposed approach does exactly that, using gradient information to inform a proposal over dimensions which leads to more efficient sampling.

\subsection{Locally-Informed Proposals}
\label{sec:locallyinformed}


A good Metropolis-Hastings proposal needs to balance the increase in the likelihood of the proposed point $x'$ with the decrease in the probability of the reverse proposal $q(x | x')$. A natural strategy for increasing the likelihood is to use locally-informed proposals:
\begin{align}
    q_{\tau}(x' | x) \propto e^{\frac{1}{\tau}(f(x') - f(x))}\mathbf{1}(x' \in H(x)).
    \label{eq:lb_prop2}
\end{align}
where $H(x)$ is the Hamming ball of some size around $x$ and $\tau > 0$ is a temperature parameter.
In this case, the proposal is simply a tempered Softmax over ${f(x') - f(x)}$ for all $x' \in H(x)$. If the temperature $\tau$ is too low, this proposal will aggressively optimize the local likelihood increase from $x \to x'$, but possibly collapse the reverse proposal probability $q_{\tau}(x | x')$. If the temperature $\tau$ is too high, this proposal may increase the reverse proposal probability $q_{\tau}(x | x')$, but ignore the local likelihood increase.

The temperature that balances both of these terms is $\tau = 2$ \citep{umrigar1993accmet, zanella2020informed}. We include a derivation of this fact in Appendix \ref{app:balancing}. In fact, \citet{zanella2020informed} showed that  $q_{2}(x' | x)$ is in the optimal subclass of locally-informed proposals. \citet{zanella2020informed} also demonstrated that this can lead to large improvements in empirical performance per sampling step compared to other generic samplers like Gibbs and the Hamming-Ball sampler~\citep{titsias2017hamming}.

Unfortunately, while powerful, these locally-informed proposals requires us to compute $f(x') - f(x)$ for every $x' \in H(x)$.
For $D$-dimensional data and a Hamming window size of 1, this requires $\mathcal{O}(D)$ evaluations of $f$ which can become prohibitive as $D$ grows. Our proposed approach reduces this to $\mathcal{O}(1)$ evaluations while incurring a minimal decrease in the efficiency of each sampling step.



\section{Searching for Structure}
For some distributions, structure exists which enables the local differences $f(x') - f(x)$ to be computed efficiently. This is not true in general, but even in settings where it is, bespoke derivations and implementations are required. Ideally, we could identify a structure that is ubiquitous across many interesting discrete distributions, can be exploited in a generic and generalizable way, and can allow us to accurately estimate the local differences. 

To find such structure, we examine the functional form of the unnormalized log-probability for some common, diverse families of discrete distributions in Table \ref{tab:disc_lll}.
\begin{table}[h!]

    \centering
    \begin{tabular}{l|l}
        \toprule
        Distribution & $\log p(x) + \log Z$ \\ 
        \midrule\midrule
        Categorical & $x^T \theta$ \\
         \midrule
        Poisson\tablefootnote{While we only present results in this paper for finite-dimensional discrete distributions, Gibbs-With-Gradients can be easily applied to (infinite) integer-valued distributions using a proposal window of $\{x-1, x + 1\}$.} & $x\log \lambda - \log \Gamma(x + 1) $\\
         \midrule
        HMM & $\sum_{t=1}^T x_{t+1}^T A x_t - \frac{(w^Tx - y)^2}{2\sigma^2}$ \\
         \midrule
        RBM  & $\sum_i \text{softplus}(W x + b)_i + c^Tx$ \\
         \midrule
        Ising & $x^T W x + b^Tx$ \\
         \midrule
    Potts  & $\sum_{i=1}^L h_i^T x_i + \sum_{i,j = 1}^L x_i^T J_{ij} x_j$\\
     \midrule
    Deep EBM & $f_\theta(x)$\\
        \bottomrule
    \end{tabular}
    \caption{Unnormalized log-likelihoods of common discrete distributions. All are differentiable with respect to $x$.}
    \label{tab:disc_lll}
\end{table}
%
%
%

The formulas in Table \ref{tab:disc_lll} are not only the standard way these distributions are written down, but they are also the standard way these distributions are implemented 
in probabilisitic programming frameworks.

The key insight here is that these are all continuous, differentiable functions accepting real-valued inputs, even though they are evaluated only on a discrete subset of their domain.
We propose to exploit this structure and that gradients, in the form of Taylor-series approximations, can be used to efficiently estimate likelihood ratios between a given input $x$ and other discrete states $x'$.


When we are dealing with $D$-dimensional binary data, we can estimate the likelihood ratios of flipping each bit with 
\begin{align}
    \tilde{d}(x) = -(2x - 1)\odot \nabla_x f(x)
    \label{eq:diff_fn_bin}
\end{align}
where $\tilde{d}(x)_i\approx f(x_{-i}) - f(x)$ and $x_{-i}$ is $x$ with the $i$-th bit flipped. If we are dealing with $D$-dimensional categorical data we can estimate a similar quantity
\begin{align}
    \tilde{d}(x)_{ij} = \nabla_x f(x)_{ij} - x_i^T \nabla_x f(x)_i
    \label{eq:diff_fn_cat}
\end{align}
where $\tilde{d}(x)_{ij}$ approximates the log-likelihood ratio of flipping the $i$-th dimension of $x$ from its current value to the value $j$.
Similar first-order approximations can easily be derived for larger window sizes as well with linear operators applied to the gradient of the log-probability function. 

We note, there may be many different underlying continuous functions which correspond to a given discrete distribution. We later present theory (Theorem 1) which can guide this choice in settings where there is any ambiguity.

\section{Gibbs With Gradients}

We now present our main algorithm. We use a Taylor-series (Equations~\ref{eq:diff_fn_bin}, \ref{eq:diff_fn_cat}) to approximate the likelihood ratios within a local window of a point $x$. We use these estimated likelihood ratios to produce an approximation
\begin{align}
     q^\nabla(x' | x) \propto e^{\frac{\tilde{d}(x)}{2}}\mathbf{1}(x' \in H(x))
    \label{eq:lb_est}
\end{align}
to $q_2(x' | x)$ of Equation~\ref{eq:lb_prop2}, which we use in the standard Metropolis-Hastings algorithm. 

Our experiments focus on a simple and fast instance of this approach which only considers local moves inside of a Hamming window of size 1. 
For binary data, these window-1 proposals have an even simpler form since all $x'\in H(x)$ differ by only dimension.
Proposing a move from $x$ to $x'$ is equivalent to choosing which dimension $i$ to change. We can sample this from a categorical distribution over $D$ choices:
\begin{align}
q(i|x) = \text{Categorical}\left(\text{Softmax}\left(\frac{\tilde{d}(x)}{2}\right)\right)
\end{align}
Thus when $x$ binary, to sample from $q^\nabla(x' | x)$, we simply sample which dimension to change $i \sim q(i|x)$, and then deterministically set $x' = \textnormal{\texttt{flipdim}}(x, i)$.
In this case, when $x$ and $x'$ differ only in dimension $i$, we have $q^\nabla(x' | x) = q(i|x)$ and $q^\nabla(x | x') = q(i|x')$.

Because of the relationship to Adaptive Gibbs, we call our sampler Gibbs-With-Gradients (GWG). Pseudo-code describing our sampler can be found in Algorithm~\ref{alg:example}. 
\begin{algorithm}[h]
  \caption{Gibbs With Gradients} 
  \label{alg:example}
\begin{algorithmic}
  \STATE {\bfseries Input:} unnormalized log-prob $f(\cdot)$, current sample $x$
  \STATE Compute $\tilde{d}(x)$ \COMMENT{Eq.~\ref{eq:diff_fn_bin} if binary, Eq.~\ref{eq:diff_fn_cat} if categorical.}
  \STATE Compute $q(i|x) = \text{Categorical}\left(\text{Softmax}\left(\frac{\tilde{d}(x)}{2}\right)\right)$
  \STATE Sample $i \sim q(i|x)$
  \STATE $x' = \textnormal{\texttt{flipdim}}(x, i)$
  \STATE Compute $q(i|x') = \text{Categorical}\left(\text{Softmax}\left(\frac{\tilde{d}(x')}{2}\right)\right)$
  \STATE Accept with probability:
  \begin{align}
      \min\left(\exp(f(x') - f(x))\frac{q(i|x')}{q(i|x)}, 1\right)\nonumber
  \end{align}
\end{algorithmic}
\end{algorithm}
In the categorical data setting, the proposal must choose not only which dimension to change, but also to what value. Thus, $q(i|x)$ in this setting is a $D(K-1)$-way Softmax.

We describe some simple extensions in Appendix \ref{app:extensions} and code to replicate our experiments is available \href{https://github.com/wgrathwohl/GWG_release}{here}.

\subsection{Analyzing Approximations}
\citet{zanella2020informed} proved that ``locally-balanced'' proposals, like $q_2(x' | x)$ in Equation \ref{eq:lb_prop2}, are the optimal locally-informed proposals for Metropolis-Hastings. In this section we show that, under smoothness assumptions on $f$, our methods are within a constant factor of $q_2(x' | x)$  in terms of asymptotic efficiency.

To understand the asymptotic efficiency of MCMC transition kernels, we can study the asymptotic variance and spectral gap of the kernel. The asymptotic variance is defined as
\begin{align}
    \text{var}_p(h, Q) = \lim_{T\rightarrow \infty} \frac{1}{T}\text{var}\left(\sum_{t=1}^Th(x_t)\right)
    \label{eq:ass_var}
\end{align}
where $h: \mathcal{X} \rightarrow R$ is a scalar-valued function, $Q$ is a $p$-stationary Markov transition kernel, and ${X_1 \sim p(x)}$. The spectral gap is defined as
\begin{align}
    \text{Gap}(Q) = 1 - \lambda_2
    \label{eq:thigh_gap}
\end{align}
where $\lambda_2$ is the second largest eigenvalue of the transition probability matrix of $Q$. Both of these quantities measure the asymptotic efficiency of $Q$. The asymptotic variances measures the additional variance incurred when using sequential samples from $Q$ to estimate $E_p[h(x)]$. For transition probability matrices with non-negative eigenvalues, the spectral gap is related to the mixing time, with larger values corresponding to faster mixing chains~\citep{levin2017markov}.

Since our method approximates $q_2(x' | x)$, we should expect some decrease in efficiency. We characterize this decrease in terms of the asymptotic variance and spectral gap, under the assumption of Lipschitz continuity of $\nabla_x f(x)$. In particular, we show that the decrease is a constant factor that depends on the Lipschitz constant of $\nabla_x f(x)$ and the window size of our proposal. 


\begin{theorem}
Let $Q(x', x)$ and $Q^\nabla(x', x)$ be the Markov transition kernels given by the Metropolis-Hastings algorithm using the locally balanced proposal $q_2(x'|x)$ and our approximation $q^\nabla(x'|x)$. 
Let $f$ be an $L$-smooth log-probability function and $p(x) = \frac{\exp(f(x))}{Z}$. Then it holds
\begin{enumerate}[label=(\alph*)]
\item $\text{var}_p(h, Q^\nabla) \leq \frac{\text{var}_p\left(h, Q\right)}{c} + \frac{1-c}{c}\cdot \text{var}_p(h)$
\item $\text{Gap}(Q^\nabla) \geq c\cdot  \text{Gap}\left(Q\right)$
\end{enumerate}
where $c = e^{-\frac{1}{2}LD_H^2}$ and $D_H = \sup_{x'\in H(x)} || x - x' ||$. 
\end{theorem}

A proof can be found in Appendix \ref{app:proof}. This roughly states that $Q^\nabla(x',x)$ is no less than $c$-times as efficient than $Q(x', x)$ per step for estimating expectations.
As expected, our approach matches the efficiency of the target proposal when the Taylor-series approximation is accurate. We can derive from Theorem 1 that when considering which functional representation of our distribution to choose, we should choose the representation whose gradient has the smallest Lipschitz constant. 

\paragraph{An example:} Consider an Ising model on a cyclic 2D lattice. This model has log-probability function ${f(x) =  \theta \cdot x^T J x - \log Z}$
where $J$ is the binary adjacency matrix of the lattice, $\theta$ is the connectivity strength and $Z$ is the unknown normalizing constant. We can see the gradient is $\nabla_x f(x) = 2\theta \cdot J x$ and can bound $L \leq 2\sigma(J) \theta = 8\theta$.

For $\theta = .1$ and a Hamming window of size 1, this gives $c = .67$, regardless of $D$.
Since a single evaluation of $f$ has an $\mathcal{O}(D^2)$ cost, it costs $\mathcal{O}(D^3)$ to compute\footnote{The local difference function of an Ising model can be computed more efficiently.  However, this requires a bespoke derivation and implementation, and is not possible for general pmfs, such as those parameterized by neural networks.}
$q_2(x'|x)$.
Compared to the exact local-differences proposal, the Gibbs-With-Gradients proposal $q^\nabla(x'|x)$ incurs at most a constant loss in sampling efficiency \emph{per-iteration} but gives a $\mathcal{O}(D)$ increase in speed.

\section{Relationship to Continuous Relaxations}

Why hasn't this relatively simple approach been proposed before?
A number of prior works have used gradient information for discrete sampling.
Instead of using gradients to inform discrete updates directly, these methods transport the problem into a continuous relaxation, perform updates there, and transform back after sampling.
This approach incurs many of the pitfalls of continuous sampling without providing the scalability.
We find these methods are not competitive with Gibbs-With-Gradients in high dimensions.


In more detail, these methods use the discrete target distribution to create a related continuous distribution (relaxation) whose samples can be transformed to samples from the target distribution. They then apply gradient-based sampling methods such as Stein Variational Gradient Descent~\citep{liu2016stein} or Hamiltonian Monte-Carlo (HMC)~\citep{neal2011mcmc} to the new distribution. Examples of such methods are the recently proposed Discrete-SVGD (D-SVGD)~\citep{han2020stein} and Discontinuous HMC~\citep{nishimura2017discontinuous}. 

A key challenge of these approaches is that the relaxed distribution can be arbitrarily difficult to sample from, highly multi-modal and require small step-sizes. Further, metrics of MCMC performance and mixing in the relaxed space may not indicate performance and mixing in the discrete space. These methods also require the tuning of many additional hyper-parameters such as step-size, momentum, and the temperature of the relaxation. In contrast, our approach operates directly in discrete space, and has no hyper-parameters.

\begin{figure}[h]
    \centering
    \includegraphics[height=.28\textwidth]{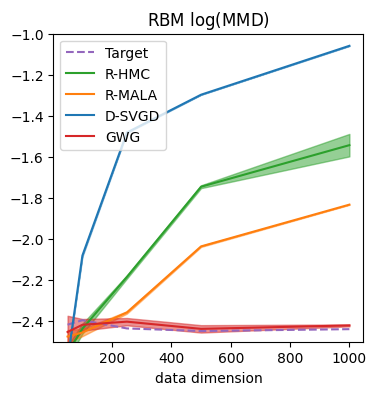}
    \includegraphics[height=.27\textwidth]{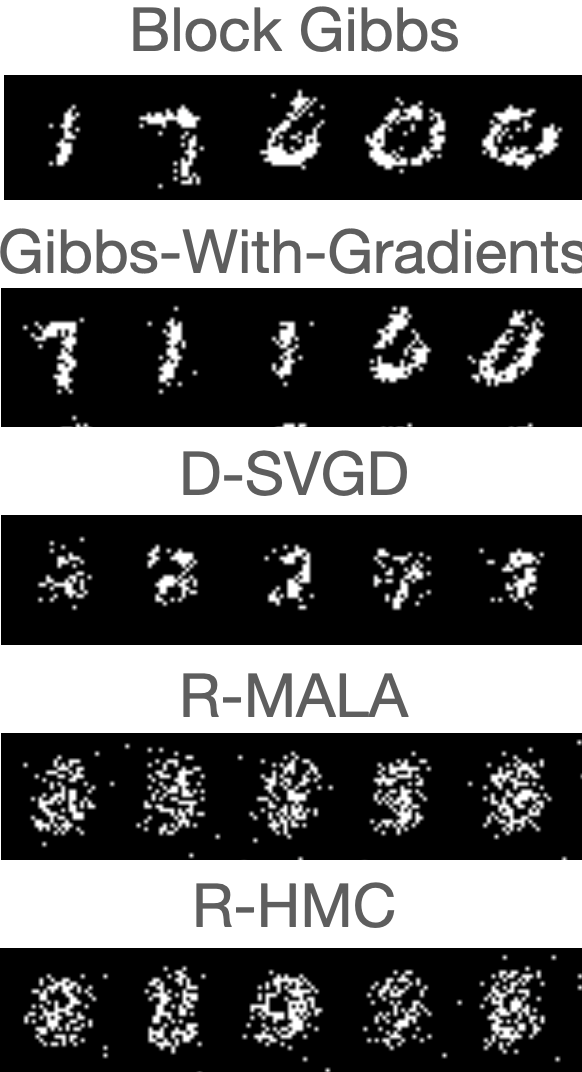}
    \caption{Comparison to gradient-based samplers with continuous relaxations. 
    GWG, D-SVGD, R-HMC, and R-MALA refer to Gibbs-With-Gradients, Discrete SVGD, Relaxed HMC and Relaxed MALA, respectively. Left: Log-MMD (lower is better) between true samples and generated samples for RBMs of increasing dimension (over 3 runs). ``Target'' is log-MMD between two sets of Block-Gibbs samples. Right: Visualized samples from an RBM trained on MNIST.}
    \label{fig:relax_rbm}
\end{figure}

Figure~\ref{fig:relax_rbm} compares these approaches on the task of sampling from restricted Boltzmann machines (RBMs) of up to 1000 dimensions.
We compare to D-SVGD and two relaxation-based baselines derived from the Metropolis-Adjusted Langevin Algorithm~\citep{besag1994comments} and HMC.
We compare the log-MMD between generated samples and ``ground-truth'' samples generated with Block-Gibbs.
We also display samples from an MNIST-trained model.
In contrast to all three baselines, our approach does not degrade with dimension.
Additional results, details, and discussion can be found in Appendix~\ref{app:relax}. These relaxation-based approaches do not scale beyond 200 dimensions, so we do not compare to them in our main experimental results section

\section{Sampling From EBMs}
To demonstrate the benefits and generality of our proposed approach to sampling, we present results sampling from 3 distinct and challenging distributions; Restricted Boltzmann Machines, Lattice Ising models, and Factorial Hidden Markov Models. Each is evaluated differently based on the properties of the distribution. We compare our sampler, Gibbs-With-Gradients, against standard Gibbs sampling and the Hamming Ball Sampler~\citep{titsias2017hamming} -- two generic approaches for discrete sampling. When available, we also compare with samplers which exploit known structure in the distribution of interest.

In the following, Gibbs-$X$ refers to Gibbs sampling with a block-size of $X$, and HB-$X$-$Y$ refers to the Hamming Ball sampler with a block size of $X$ and a hamming ball size of $Y$, and GWG refers to Gibbs-With-Gradients. Gibbs-1 is the fastest sampler tested. In our current implementation, we find Gibbs-2, HB-10-1, and GWG have approximately 1.6, 6.6, 2.1 times the cost of Gibbs-1 per step, respectively. Thus the run-time of GWG is most comparable to Gibbs-2. 

\begin{figure}[h]
    \centering
    \includegraphics[height=.25\textwidth]{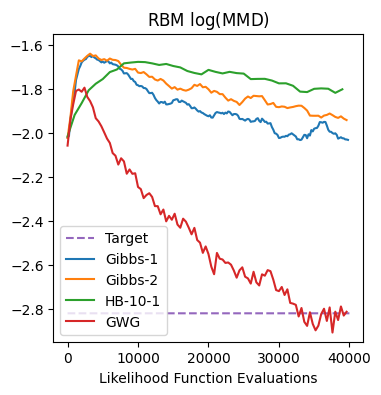}
     \includegraphics[height=.25\textwidth]{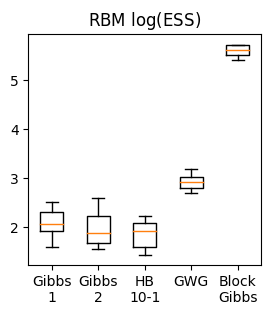}
    \caption{RBM sampling results. Left: Log-MMD of samples over steps (lower is better). ``Target'' is Log-MMD between two sets of Block-Gibbs samples. Right: Log-ESS of various samplers after 10,000 steps. Gibbs-With-Gradients matches Block-Gibbs in MMD and outperforms unstructured baselines in ESS.
    }
    \label{fig:samp_rbm}
\end{figure}

\paragraph{Restricted Boltzmann Machines} are unnormalized latent-variable models defined as:
\begin{align}
    \log p(x) = \log (1 + \exp(Wx + c)) + b^Tx - \log Z
\end{align}
where $\{W, b, c\}$ define its parameters and $x \in \{0, 1\}^D$. We train an RBM with 500 hidden units on the MNIST dataset using contrastive divergence~\cite{hinton2002training}. We generate samples with various MCMC samplers and compare them in two ways. First, using the Maximum Mean Discrepancy (MMD)~\citep{gretton2012kernel} between a set of samples from each sampler and a set of ``ground-truth'' samples generated using the structured Block-Gibbs sampler available to RBMs (see Appendix \ref{app:rbm} for details). Next, we report the Effective Sample Size (ESS) of a summary statistic over sampling trajectories. Results can be seen in Figure \ref{fig:samp_rbm}.

We see on the left that GWG matches the structured Block-Gibbs sampler in MMD (``Target'' in the Figure), while the other samplers do not. On the right we see that the effective sample size of GWG is notably above the baselines and is approximately halfway between the baselines and the Block-Gibbs sampler (in log-space). We note, Block-Gibbs can update all 784 dimensions in each iterations. GWG and Gibbs-1 can update 1 and Gibbs-2 and Hamming Ball can update 2 dimensions per iteration. 

\paragraph{Lattice Ising Models} are models for binary data defined by
\begin{align}
    \log p(x) =  \theta \cdot x^T J x - \log Z
\end{align}
where $\theta$ is the connectivity strength and $J$ is the binary adjacency matrix, here restricted to be a 2D cyclic lattice. This model was originally proposed to model the spin magnetic particles~\citep{ising1924beitrag}. 
We sample from models with increasing dimension and connectivity strength. We evaluate using Effective Sample Size (ESS)\footnote{Computed using \href{https://www.tensorflow.org/probability/api_docs/python/tfp/mcmc/effective_sample_size}{Tensorflow Probability}} of a summary statistic (full details in Appendix \ref{app:ising}). Results can be seen in Figure \ref{fig:samp_ising}.
\begin{figure}[h]
    \centering
    \includegraphics[height=.23\textwidth]{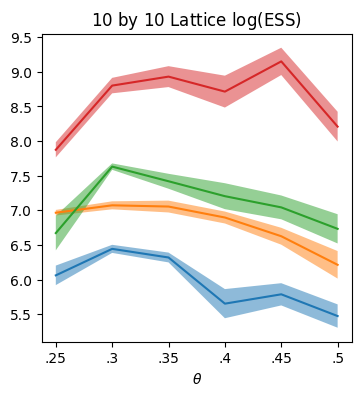}
     \includegraphics[height=.23\textwidth]{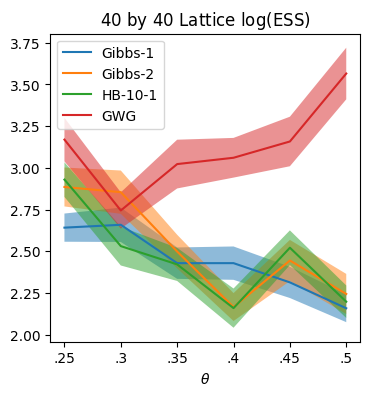}
    \caption{Ising model sampling results. The y-axis shows log-ESS over 100,000 samples steps. Left: 10x10 lattice, right: 40x40 lattice. We see Gibbs-With-Gradients (GWG) outperforms in most settings.}
    \label{fig:samp_ising}
\end{figure}

We see Gibbs-With-Gradients provides a notable increase in ESS for Ising models with higher connectivity. These models are harder to sample from as their dimensions are more correlated.  

\begin{figure}[h]
    \centering
    \includegraphics[height=.23\textwidth]{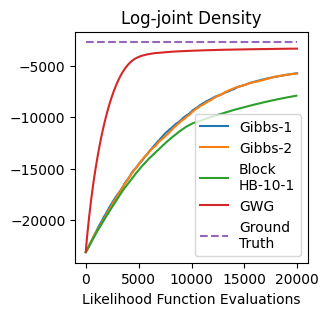}
     \includegraphics[height=.23\textwidth]{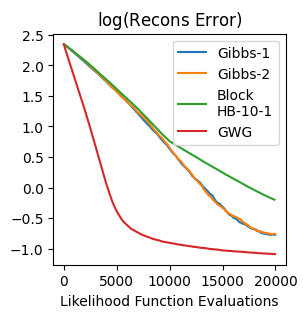}
    \caption{Factorial Hidden Markov Model results.     ``Block-HB'' refers to the block-structured hamming ball sampler. Left, log-joint density and right, mean log-reconstruction error. GWG performs best in both evaluations, outperforming the Hamming Ball sampler which exploits model structure.}
    \label{fig:samp_fhmm}
\end{figure}

\paragraph{Factorial Hidden Markov Models} (FHMM) are latent-variable time-series models, similar to HMMs but their hidden state consists of distinct, independent factors. The continuous data $y \in R^L$ of length $L$ is generated by the binary hidden state $x \in \{0, 1\}^{L\times K}$ with $K$ factors as  
\begin{align}
    p(x, y) &= p(y|x)p(x)\nonumber\\
    p(y|x) &= \prod_{t=1}^L \mathcal{N}(y_t; Wx_t + b, \sigma^2)\nonumber\\
    p(x) &= p(x_1)\prod_{t=2}^L p(x_t|x_{t-1})
\end{align}
We create a random FHMM with 1000 time-steps and a 10-dimensional hidden state and then draw samples $y$. We generate posterior samples $p(x|y)$ and evaluate our samplers using reconstruction error and joint likelihood. Full model description and experimental details can be found in Appendix \ref{app:fhmm} and results can be seen in Figure \ref{fig:samp_fhmm}.

In this setting, the Hamming-Ball sampler exploits known structure in the problem. Each block chosen by the sampler consists of the 10-dimensional latent state $x_t$, as opposed to 10 random dimensions. Thus, the Hamming-Ball sampler in this setting is a stronger baseline. Despite this, we find Gibbs-With-Gradients notably outperforms the baseline samplers.




\section{Training EBMs}
Training EBMs is a challenging task. Computing the likelihood for Maximum Likelihood inference requires computation of the normalizing constant ${Z = \sum_x e^{f_\theta(x)}}$ which is typically intractable. Thankfully, the gradient of the log-likelihood can be more easily expressed as:
\begin{align}
    \nabla_\theta \log p(x) = \nabla_\theta f_\theta(x) - E_{p(x)}[\nabla_\theta f_\theta(x)] 
    \label{eq:ml_grad_est}
\end{align}
therefore, if samples can be drawn from $p(x)$, then an unbiased gradient estimator can be derived. We can approximate this estimator using MCMC. When a slow-mixing MCMC sampler is used to draw these samples, we obtain biased gradient estimates and this leads to sub-par learning. Improvements in MCMC can then lead to improvements in parameter inference for unnormalized models. We explore how Gibbs-With-Gradients can be applied to parameter inference for
some classical discrete EBMs.

\subsection{Training Ising models on generated data}
We generate Ising models with different sparse graph structures; a 2D cyclic lattice and a random Erdos-Renyi graph. We generate training data with a long-run Gibbs chain and train models using Persistent Contrastive Divergence~\citep{tieleman2008training}
with an $\ell1$ penalty to encourage sparsity. We evaluate our models using the Root-Mean-Squared-Error (RMSE) between the inferred connectivity matrix $\hat{J}$ and the true matrix $J$.

\begin{figure}[h]
    \centering
      \includegraphics[height=.22\textwidth]{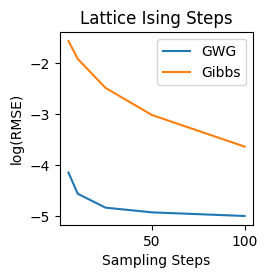}
     \includegraphics[height=.22\textwidth]{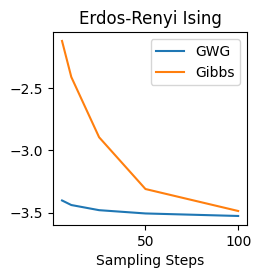}
    \caption{Training Ising models with increasing MCMC steps.
    Left: Lattice Ising (dim = 625, $\theta = .25$). Right: Erdos-Renyi Ising. Values are $\log$(RMSE) between the learned and true $J$. GWG leads to better solutions with lower computational cost. }
    \label{fig:tr_ip}
\end{figure}

Full experimental details and additional results can be found in Appendix \ref{app:tr_ip}, \ref{app:tr_ip_add} and results can be seen in Figure \ref{fig:tr_ip}. In all settings, GWG greatly outperforms Gibbs sampling.
This allows for much faster training than standard Gibbs while recovering higher-quality solutions.

\subsection{Protein Coupling Prediction with Potts Models}

Proteins are defined by a discrete sequence of 20 amino acids $x\in \{1,\ldots, 20\}^D$ where $D$ is the length of the protein. 
The Potts model has long been a popular approach for modelling the evolutionary distribution of protein sequences~\citep{lapedes1999correlated}. The model takes the form
\begin{align}
    \log p(x) = \sum_{i=1}^D h_i^T x_i + \sum_{i,j = 1}^D x_i^T J_{ij} x_j - \log Z
\end{align}
where $x_i$ is a one-hot encoding of the $i$-th amino acid in $x$, $J \in R^{\{D\times D\times 20 \times 20\}}$ and $h\in R^{\{D \times 20\}}$ are the model's parameters and $Z$ is the model's normalizing constant. The Potts model's likelihood is the sum of pairwise interactions. \citet{marks2011protein} demonstrated that the strength of these interactions can correspond to whether or not two amino acids touch when the protein folds. These inferred contacts can then be used to infer the 3D structure of the protein. 

Since the Potts model is unnormalized, maximum likelihood learning is difficult, and $\ell1$-regularized Pseudo-likelihood Maximization (PLM)~\citep{besag1975statistical} is used to train the model. Recently \citet{ingraham2017variational} found that improved contact prediction could be achieved with MCMC-based maximum likelihood learning. Unfortunately, due to the limitations of discrete MCMC samplers, their study was restricted to small proteins (less than 50 amino acids).

GWG allows these performance improvements to scale to large proteins as well. We train Potts models on 2 large proteins: OPSD\_BOVIN, and CADH1\_HUMAN. We train using PCD where samples are generated with GWG and Gibbs. We run PLM as a baseline. These proteins are much larger than those studied in \citet{ingraham2017variational} with OPSD\_BOVIN, and CADH1\_HUMAN having 276, and 526 amino acids, respectively\footnote{After standard data pre-processing as in \citet{ingraham2017variational}}. 

We predict couplings using the $J$ parameter of our models. We compute a ``coupling-strength'' for each pair of amino-acids as $||J_{ij}||_2$ which gives a measure of how much indices $i$ and $j$ interact to determine the fitness of a protein. We sort index pairs by their coupling strength and compare the highest scoring pairs with known contacts in the proteins. 

\begin{figure}[h]
    \centering
    \includegraphics[width=.23\textwidth]{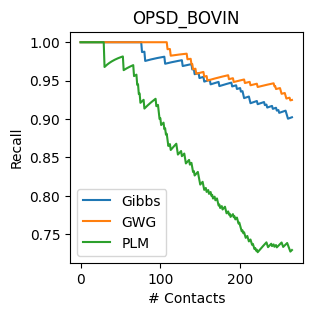}
     \includegraphics[width=.23\textwidth]{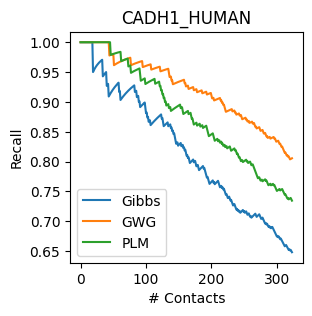}
    \caption{Recall Curves for contact prediction with Potts models.
    Gibbs-With-Gradients leads to higher recall.}
    \label{fig:tr_prot}
\end{figure}

Full experimental details and additional results can be found in Appendix \ref{app:tr_prot} and results can be seen in Figure \ref{fig:tr_prot}. For the smaller protein, Gibbs sampling outperforms PLM but for the larger protein, the slow-mixing of the sampler causes the performance to drop below that of PLM. Despite the increased size, Gibbs-With-Gradients performs the best. 

\begin{table*}[h!]

    \centering
    \begin{tabular}{c|l|cccccc}
        \toprule
        \multirow{2}{*}{Data Type} & \multirow{2}{*}{Dataset} & VAE  & VAE    & EBM & EBM & \multirow{2}{*}{RBM} & \multirow{2}{*}{DBN}  \\
                  &         & (MLP) & (Conv) &   (GWG)  &   (Gibbs)       &     &    \\
        \midrule
        \multirow{2}{*}{Binary} & Static MNIST  & −86.05  & −82.41 & \textbf{−80.01} & -117.17 & -86.39 & -85.67  \\
        & Dynamic MNIST  & −82.42 & \textbf{−80.40} & −80.51 & -121.19 & --- & --- \\
        \multirow{2}{*}{(log-likelihood $\uparrow$)} & Omniglot   & −103.52  & −97.65 & \textbf{-94.72} & -142.06 & -100.47 & -100.78 \\
        & Caltech Silhouettes & −112.08 & −106.35 & \textbf{−96.20} & -163.50 & --- & --- \\
        \midrule
        Categorical & Frey Faces   & 4.61 & \textbf{4.49} &  4.65 & --- & --- & --- \\
        (bits/dim $\downarrow$) & Histopathology   & 5.82 & 5.59  & \textbf{5.08} & ---   & --- & --- \\
        \bottomrule
    \end{tabular}
    \caption{Test-set log-likelihoods for models trained on discrete image datasets. RBM and DBN results are taken from \citet{burda2015accurate}, VAE results taken from \citet{tomczak2018vae}.}
    \label{tab:ebm_ll}
\end{table*}
 
\section{Deep EBMs for Discrete Data}
\label{sec:deep}
Deep Energy-Based Models have rapidly gained popularity for generative modeling. These models take the form
\begin{align}
    \log p(x) = f_\theta(x) - \log Z
    \label{eq:deep_ebm}
\end{align}
where $f_\theta: R^D \rightarrow R$ is a deep neural network. 
The recent success of these models can be attributed to a few advancements including; the use of tempered Langevin samplers~\citep{nijkamp2020anatomy} and large persistent chains~\citep{du2019implicit}. This has enabled EBMs to become a competitive approach for image-generation~\citep{song2019generative}, adversarial robustness~\citep{grathwohl2019your, hill2020stochastic}, semi-supervised learning~\citep{song2018learning, grathwohl2020no}
and many other problems. 

These advances rely on gradient-based sampling which requires continuous data. Thus, these scalable methods cannot be applied towards training deep EBMs on discrete data. We explore how Gibbs-With-Gradients can enable the training of deep EBMs on high-dimensional binary and categorical data. To our knowledge, models of this form have not be successfully trained on such data in the past. 
We train deep EBMs parameterized by Residual Networks~\citep{he2016deep} on small binary and continuous image datasets using PCD~\citep{tieleman2008training} with a replay buffer as in \citet{du2019implicit, grathwohl2019your}. The continuous images were treated as 1-of-256 categorical data.




PCD training is very sensitive to the choice of MCMC sampler. 
As an initial experiment, we attempted to train these models using standard Gibbs but found that the sampler was too slow to enable stable training within a reasonable compute budget.
On the binary data we needed to train with \textbf{800} Gibbs sampling steps per training iteration. All models we trained with fewer steps quickly diverged. GWG required only 40. This made training with Gibbs 9.52x slower than GWG.
For a fair comparison, the Gibbs results in Table \ref{tab:ebm_ll} were trained for an equal amount of wall-clock time as the GWG models. 

For the categorical data, we could not train models with Gibbs sampling. Each step of Gibbs requires us to evaluate the energy function 256 (for each possible pixel value) times. GWG requires 2 function evaluations. Thus the amount of compute per iteration for Gibbs is 128x greater than GWG. Further, to make Gibbs train stably, we would need to use many more steps, as with the binary data. This would give roughly a \textbf{870x} increase in run-time. Therefore, training a model of this form with Gibbs is simply not feasible.

\begin{figure}[h]
    \centering
    \includegraphics[width=.23\textwidth]{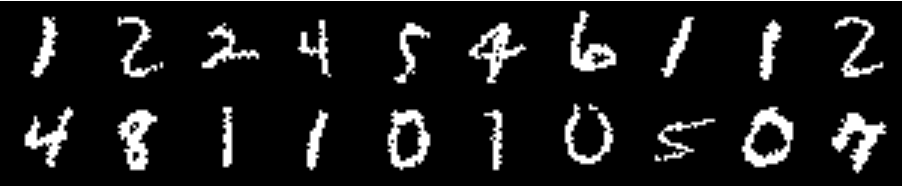}
    \includegraphics[width=.23\textwidth]{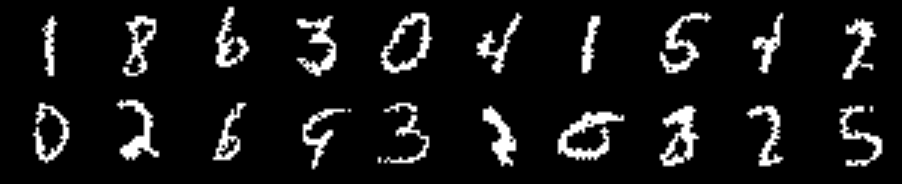}
    \includegraphics[width=.23\textwidth]{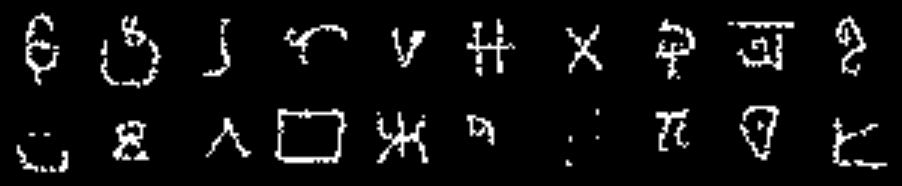}
    \includegraphics[width=.23\textwidth]{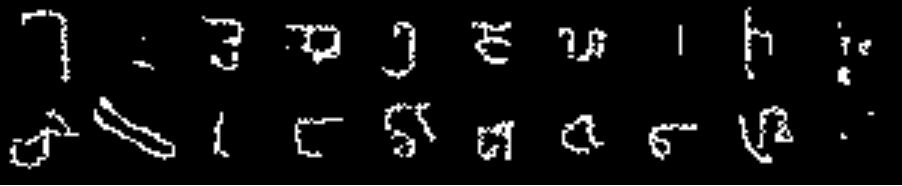}
    \includegraphics[width=.23\textwidth]{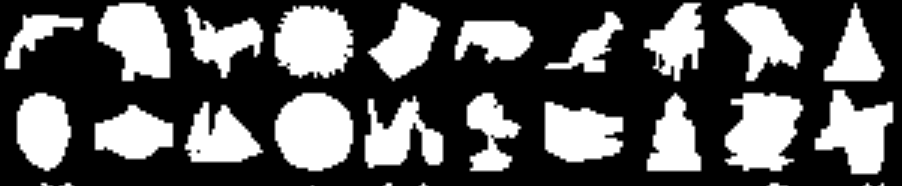}
    \includegraphics[width=.23\textwidth]{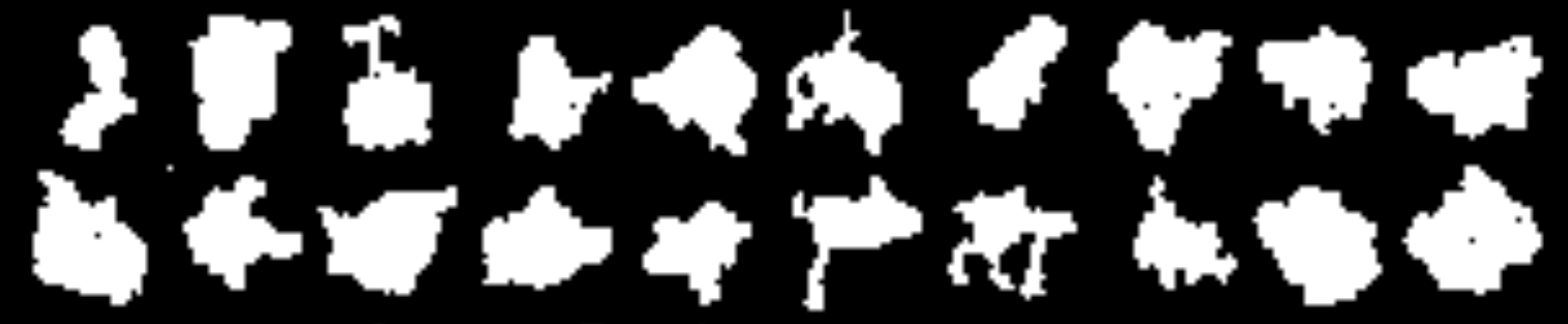}
     \includegraphics[width=.23\textwidth]{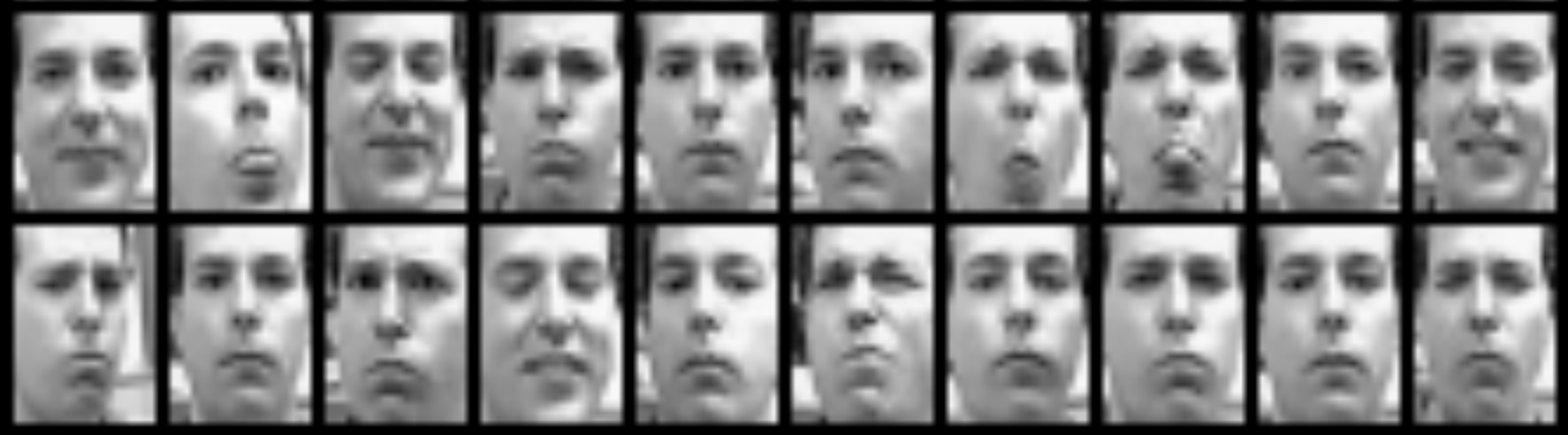}
    \includegraphics[width=.23\textwidth]{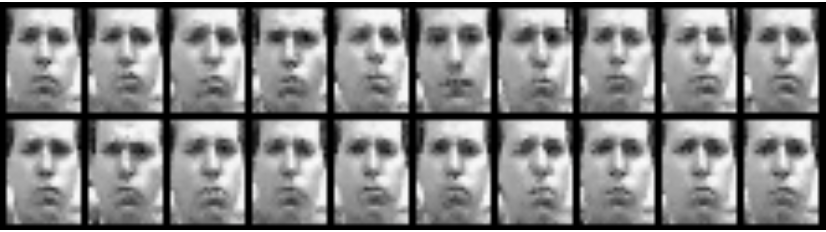}
    \includegraphics[height=.047\textwidth]{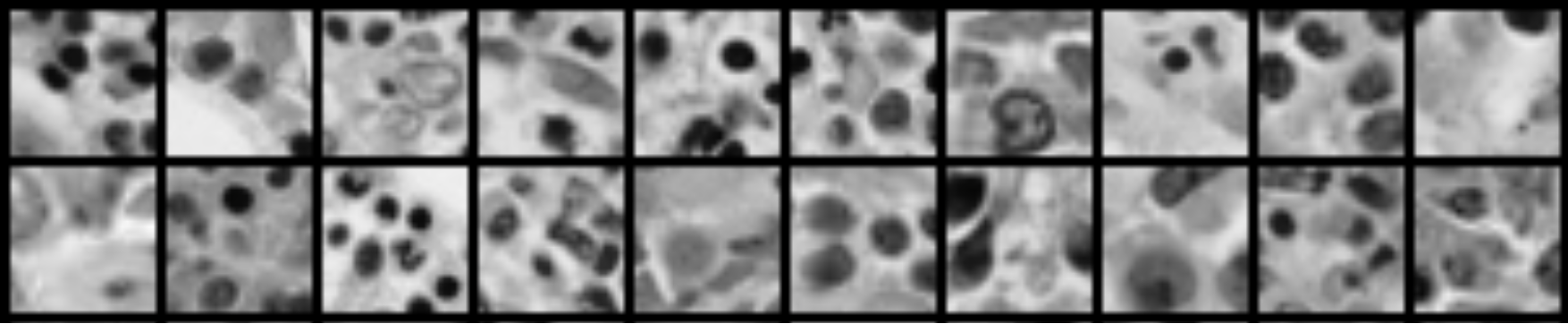}
    \includegraphics[height=.047\textwidth]{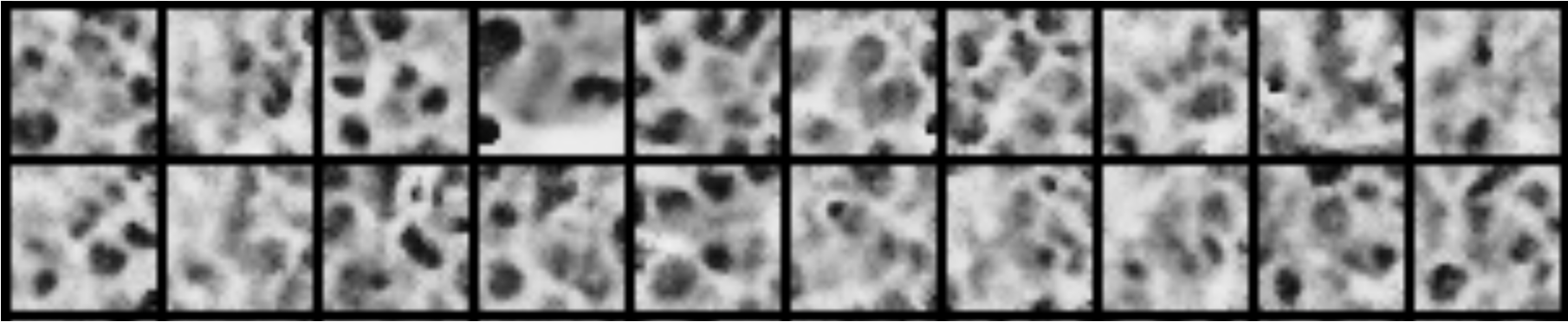}
    \caption{Left: data. Right: Samples from ResNet EBM. Samples generated with annealed Markov chain using 300,000 Gibbs-With-Gradients steps. Top to bottom: MNIST, omniglot, Caltech Silhouettes, Frey Faces,  Histopathology.}
    \label{fig:resnet_ebm}
\end{figure}

Full experimental details can be found in Appendix \ref{app:deep_ebm}. We present long-run samples from our trained models in Figure \ref{fig:resnet_ebm} and test-set likelihoods in Table \ref{tab:ebm_ll}. Likelihoods are estimated using Annealed Importance Sampling~\citep{neal2001annealed}. 
We compare the performance of our models to Variational Autoencoders~\citep{kingma2013auto} and two other EBMs; an RBM and a Deep Belief Network (DBN)~\citep{hinton2009deep}. On most datasets, our Resnet EBM outperforms the other two EBMs and the VAEs. Our improved sampler enables deep EBMs to become a competitive approach to generative modeling on high-dimensional discrete data.

We include some preliminary results using Gibbs-With-Gradients to train EBMs for text data in Appendix \ref{app:text}.



\section{Future Directions and Conclusion}
In this work we have presented Gibbs-With-Gradients, a new approach to MCMC sampling for discrete distributions. Our approach exploits a powerful structure, gradient information, which is available to a very large class of important discrete distributions. We then use this gradient information to construct proposal distributions for Metropolis-Hastings.

We have demonstrated on a diverse set of distributions that this approach to sampling considerably outperforms baseline samplers which do not exploit known structure in the target distribution as well as many that do. Further, we find our approach outperforms prior discrete samplers which use gradient information with continuous relaxations. 

We find Gibbs-With-Gradients performs very well at sampling from deep energy-based models and allows, for the first time, unconstrained deep EBMs to be trained on discrete data and outperform other deep generative models. 

We believe there is considerable room for future work building on top of our method. We only explored samplers which modify 1 variable per proposed update. We believe considerable improvements could be made if the window size of the sampler was expanded but this would require more efficient algorithms to sample from the larger proposal. 

Next, we have shown that gradient-based approximations to the local difference function can be accurate and useful for complex discrete distributions. Local difference functions have been used in the past to generalize Score Matching~\citep{lyu2012interpretation}, and Stein Discrepancies~\citep{han2018stein}. We believe there is great potential to explore how gradient-based approximations could enable the generalization of recent deep EBM training methods based on Score Matching and Stein Discrepancies~\citep{song2019generative, grathwohl2020learning} to models of discrete data.

\section{Acknowledgements}
We would like to thank Eli Weinstein for helping us properly present our protein model results and we would like to thank Kelly Brock for help and feedback for working with the protein data. We thank Jesse Bettencourt, James Lucas, and Jacob Kelly for helpful feedback on our draft. Last, we would like to thank Jeffrey Rosenthal for providing information on related methods and prior work. 




\newpage\newpage
\bibliography{example_paper}
\bibliographystyle{icml2021}

\newpage\clearpage
\appendix
\section{Balancing Locally-Informed Proposals}
\label{app:balancing}
As discussed in Section \ref{sec:locallyinformed}, locally informed proposals need to balance the likelihood increase with the reverse proposal probability. In particular, consider proposals of the form:
\begin{align}
    q_\tau(x' | x) \propto e^{\frac{1}{\tau}(f(x') - f(x))}\mathbf{1}(x' \in H(x)).
    \label{eq:lb_prop2_app}
\end{align}
where $H(x)$ is the Hamming ball of some size around $x$ and $\tau > 0$ is a temperature parameter. Here we provide a derivation of the fact that $\tau = 2$ balances these two terms.

When we examine the acceptance rate (Equation \ref{eq:mh}) of this proposal we find
\begin{align}
  \exp(&f(x') - f(x)) \frac{q_\tau(x|x')}{q_\tau(x'|x)}\nonumber\\&= \exp(f(x') - f(x)) \frac{\exp(\frac{1}{\tau}(f(x) - f(x')) Z(x)}{\exp(\frac{1}{\tau}(f(x') - f(x)) Z(x')}\nonumber\\
    &= \exp\left(\left(1 - \frac{2}{\tau}\right)(f(x') - f(x))\right)\frac{Z(x)}{Z(x')}
\end{align}
where $Z(x) = \sum_{x'\in H(x)} \exp(\frac{1}{\tau}(f(x') - f(x))$ is the normalizing constant of the proposal. By setting $\tau = 2$, the most variable terms cancel and we are left with 
$\frac{Z(x)}{Z(x')}$. Thus, the acceptance rate is equal to the ratio of the normalizing constants of the proposal distributions. If the Hamming ball is small and the function $f$ is well behaved (i.e Lipschitz) then, since $x'$ is near $x$, $Z(x')$ will be near $Z(x)$ and the acceptance rate will be high.





\section{Proof of Theorem 1}
\label{app:proof}
Our proof follows from Theorem 2 of \citet{zanella2020informed} which states that for two $p$-reversible Markov transition kernels $Q_1(x', x)$ and $Q_2(x', x)$, if there exists $c > 0$ for all $x' \neq x$ such that $Q_1(x', x) > c \cdot Q_2(x', x)$ then 
\begin{enumerate}[label=(\alph*)]
    \item $\text{var}_p(h, Q_1) \leq \frac{\text{var}_p(h, Q_1)}{c}  + \frac{1 - c}{c} \cdot \text{var}_p(h)$
    \item $\text{Gap}(Q_1) \geq c \cdot \text{Gap}(Q_2)$
\end{enumerate}
where $\text{var}_p(h, Q)$ is the asymptotic variance defined in Equation \ref{eq:ass_var}, $\text{Gap}(Q)$ is the spectral gap defined in Equation \ref{eq:thigh_gap}, and $\text{var}_p(h)$ is the standard variance $E_p[h(x)^2] - E_p[h(x)]^2$.

Our proof proceeds by showing we can bound ${Q^\nabla(x', x) \geq c \cdot Q(x', x)}$, and the results of the theorem then follow directly from Theorem 2 of \citet{zanella2020informed}.

\subsection{Definitions}
We begin by writing down the proposal distribution of interest and their corresponding Markov transition kernels. For ease of notion we define some values
\begin{align}
    \Delta(x', x) &\coloneqq f(x') - f(x)\nonumber\\
    \nabla(x', x) &\coloneqq \nabla_x f(x)^T (x' - x)\nonumber\\
    D_H &\coloneqq \sup_{x' \in H(x)}||x' - x||\nonumber
\end{align}

We restate the target proposal for $x' \in H(x)$
\begin{align*}
    q_2(x'| x) = \frac{\exp\left(\frac{\Delta(x', x)}{2}\right)}{Z(x)}
\end{align*}
where we have defined
\begin{align*}
    Z(x) = \sum_{x' \in H(x)} \exp\left(\frac{\Delta(x', x)}{2}\right).
\end{align*}

This proposal leads to the Markov transition kernel
\begin{align*}
    Q(x', x) &= q_2(x'|x)\min\left\{1, \frac{Z(x)}{Z(x')}\right\}\nonumber\\
    &= \min\left\{\frac{\exp\left(\frac{\Delta(x', x)}{2}\right)}{Z(x)}, \frac{\exp\left(\frac{\Delta(x', x)}{2}\right)}{Z(x')}\right\}.
\end{align*}

We now restate our approximate proposal for $x' \in H(x)$
\begin{align*}
    q^\nabla(x'| x) = \frac{\exp\left(\frac{\nabla(x', x)}{2}\right)}{\tilde{Z}(x)}
\end{align*}
where we have defined
\begin{align*}
    \tilde{Z}(x) = \sum_{x' \in H(x)} \exp\left(\frac{\nabla(x', x)}{2}\right)
\end{align*}
which leads to the Markov transition kernel
\begin{align*}
    Q&^\nabla(x', x) =\nonumber\\
    &q^\nabla(x'|x)\min\left\{1, \frac{\exp \left(\Delta(x
    ', x)\right)}{\exp\left(\frac{\nabla(x', x)-\nabla(x, x')}{2}\right)}\frac{\tilde{Z}(x)}{\tilde{Z}(x')}\right\}\nonumber\\
    &= \min\left\{\frac{\exp\left(\frac{\nabla(x', x)}{2}\right)}{\tilde{Z}(x)}, \frac{\exp\left(\Delta(x',x) + \frac{\nabla(x, x')}{2}\right)}{\tilde{Z}(x')}\right\}.
\end{align*}

\subsection{Preliminaries}
It can be seen that $\nabla(x', x)$ is a first order Taylor-series approximation to $\Delta(x', x)$ and it follows directly from the Lipschitz continuity of $\nabla_x f(x)$ that 
\begin{align}
    |\nabla(x', x) - \Delta(x', x)| \leq \frac{L}{2} ||x' - x'||^2
\end{align}
and since we restrict $x' \in H(x)$ we have
\begin{align}
    -\frac{L}{2} D_H^2 \leq \nabla(x', x) - \Delta(x', x) \leq \frac{L}{2}D_H^2
\end{align}

\subsection{Normalizing Constant Bounds}
We derive upper- and lower-bounds on $\tilde{Z}(x)$ in terms of $Z(x)$.
\begin{align}
    \tilde{Z}(x) &=  \sum_{x' \in H(x)} \exp\left(\frac{\nabla(x', x)}{2}\right)\nonumber\\
    &= \sum_{x' \in H(x)} \exp\left(\frac{\Delta(x', x)}{2}\right) \exp\left(\frac{\nabla(x', x) - \Delta(x', x)}{2}\right)\nonumber\\
    &\leq \sum_{x' \in H(x)} \exp\left(\frac{\Delta(x', x)}{2}\right) \exp\left(\frac{LD_H^2}{4}\right)\nonumber\\
    &\leq \exp\left(\frac{LD_H^2}{4}\right)\sum_{x' \in H(x)} \exp\left(\frac{\Delta(x', x)}{2}\right)\nonumber\\
    &= \exp\left(\frac{LD_H^2}{4}\right)Z(x)
\end{align}

Following the same argument we can show
\begin{align}
    \tilde{Z}(x) &\geq \exp\left(\frac{-LD_H^2}{4}\right)Z(x).
\end{align}

\subsection{Inequalities of Minimums}
We show ${Q^\nabla(x', x) \geq c\cdot  Q(x', x)}$ for ${c = \exp\left(\frac{-LD_H^2}{2}\right)}$. Since both 
\begin{align*}
    Q(x', x) = \min\{a, b\}
\end{align*}
and 
\begin{align*}
    Q^\nabla(x', x) = \min\{a^\nabla, b^\nabla\}
\end{align*}
it is sufficient to show $a^\nabla \geq c\cdot a $ and $b^\nabla \geq c\cdot b$ to prove the desired result.

We begin with the $a$ terms
\begin{align}
    \frac{a^\nabla}{a} &= \frac{\exp\left(\frac{\nabla(x', x)}{2}\right)}{\tilde{Z}(x)} \frac{Z(x)}{\exp\left(\frac{\Delta(x', x)}{2}\right)}\nonumber\\
    &= \frac{Z(x)}{\tilde{Z}(x)}\exp\left(\frac{\nabla(x', x)}{2} - \frac{\Delta(x', x)}{2}\right)\nonumber\\
    &\geq \exp\left(\frac{-LD_H^2}{4}\right)\exp\left(\frac{\nabla(x', x)}{2} - \frac{\Delta(x', x)}{2}\right)\nonumber\\
     &\geq \exp\left(\frac{-LD_H^2}{4}\right)\exp\left(\frac{-LD_H^2}{4}\right)\nonumber\\
     &= \exp\left(\frac{-LD_H^2}{2}\right)
\end{align}

Now the $b$ terms
\begin{align}
    \frac{b^\nabla}{b} &=  \frac{\exp\left(\Delta(x',x) + \frac{\nabla(x, x')}{2}\right)}{\tilde{Z}(x')} \frac{Z(x')}{\exp\left(\frac{\Delta(x', x)}{2}\right)}\nonumber\\
    &= \frac{Z(x')}{\tilde{Z}(x')} \frac{\exp\left(\Delta(x',x) + \frac{\nabla(x, x')}{2}\right)}{\exp\left(\frac{\Delta(x', x)}{2}\right)}\nonumber\\
    &= \frac{Z(x')}{\tilde{Z}(x')} \exp\left(\frac{\Delta(x',x)}{2} + \frac{\nabla(x, x')}{2}\right)\nonumber\nonumber\\
    &\geq \exp\left(\frac{-LD_H^2}{4}\right) \exp\left(\frac{\Delta(x',x)}{2} + \frac{\nabla(x, x')}{2}\right)\nonumber\\
    &= \exp\left(\frac{-LD_H^2}{4}\right) \exp\left(\frac{\nabla(x, x')}{2} - \frac{\Delta(x,x')}{2}\right)\nonumber\\
    &\geq \exp\left(\frac{-LD_H^2}{4}\right) \exp\left(\frac{-LD_H^2}{4}\right)\nonumber\\
    &= \exp\left(\frac{-LD_H^2}{2}\right)
\end{align}

\subsection{Conclusions}
We have shown that ${a^\nabla \geq \exp\left(\frac{-LD_H^2}{2}\right) a}$ and ${b^\nabla \geq \exp\left(\frac{-LD_H^2}{2}\right) b}$ and therefore it holds that 
\begin{align}
    Q^\nabla(x', x) \geq \exp\left(\frac{-LD_H^2}{2}\right) Q(x', x)
\end{align}

From this, the main result follows directly from Theorem 2 of \citet{zanella2020informed}.

\section{Relationship to Relaxations}
\label{app:relax}
\citet{han2020stein} show that sampling from any discrete distribution can be transformed into sampling from a continuous distribution with a piece-wise density function. For simplicity we focus on a distribution $p(x)$ over binary data ${x \in \{0, 1\}^D}$. To do this we will create a $D$-dimensional continuous distribution $p_c(z)$ where $z \in R^D$. We must specify a base distribution $p_0(z)$ which we choose to be $\mathcal{N}(0, I)$. We must then specify a function $\Gamma(z): R^D \rightarrow \{0, 1\}^D$ which maps regions of equal mass under the base distribution to values in $\{0, 1\}^D$. A natural choice is $\Gamma(z) = \text{sign}(z)$

We then define $p_c(z)$ as
\begin{align*}
    p_c(z) = \mathcal{N}(z; 0, I)p(\Gamma(z))
\end{align*}
and it can be easily verified that generating
\begin{align*}
    z \sim p_c(z), \qquad x = \Gamma(z)
\end{align*}
will produce a sample from $p(x)$.

Thus, we have transformed a discrete sampling task into a task of sampling from a continuous distribution with a piece-wise density. \citet{han2020stein} further relax this by defining
\begin{align*}
    p^\lambda_c(x) = \mathcal{N}(z; 0, I)p(\Gamma_\lambda(z))
\end{align*}
where $\Gamma_\lambda(x)$ is a continuous approximation to $\Gamma(x)$. A natural choice for $\text{sign}(x)$ is a tempered sigmoid function
\begin{align*}
\Gamma_\lambda(x) = \frac{1}{1 + e^{-x/\lambda}}
\end{align*}
with temperature $\lambda$ which controls the smoothness of the relaxation. This is similar to the Concrete relaxation~\citep{maddison2016concrete} for binary variables. 

D-SVGD proposes to use the gradients of $\log p^\lambda_c(x)$ to produce updates for their continuous samples which are adjusted using an importance-weighted correction as proposed in \citet{han2018stein}.

We can apply this same approach to other sampling methods such as MALA and HMC.

\subsection{Relaxed MCMC}
Gradient-based MCMC samplers such as MALA or HMC consist of a proposal distribution, $q(x'|x)$, and a metropolis accept/reject step. As a baseline of comparison, we present two straight-forward applications of the above relaxation to sampling from discrete distributions. In both settings we will use the continuous, differentiable surrogate $p^\lambda_c(x)$ to generate a proposal and we will perform our Metropolis step using the piece-wise target $p_c(x)$.

\paragraph{Relaxed MALA (R-MALA):} Given a sample $x$, we sample a proposed update $x'$ from:
\begin{align}
    q(x'|x) = \mathcal{N}\left(x' ; x + \frac{\epsilon}{2}\nabla_x \log p^\lambda_c(x), \epsilon^2\right)
\end{align}
and we accept this proposal with probability
\begin{align}
    \min\left\{\frac{p_c(x')q(x|x')}{p_c(x)q(x'|x)}, 1 \right\}.
\end{align}

R-MALA has two parameters; the stepsize $\epsilon$ and the temperature of the relaxation $\lambda$. We search over $\epsilon \in \{.1, .01, .001\}$ and $\lambda \in \{2., 1., .5\}$

\paragraph{Relaxed HMC (R-HMC)} works similarly to R-MALA. Given a sample $x$ we sample an initial momentum vector $v \sim \mathcal{N}(0, M)$ where $M$ is the mass matrix. We perform $k$ steps of leapfrog integration on the relaxed Hamiltonian
\begin{align*}
    H^\lambda(x, v) = -\log p^\lambda_c(x) + \frac{1}{2} v^T M v
\end{align*}
with step-size $\epsilon$.

The proposal $x', v'$ is accepted with probability
\begin{align}
    \min\left\{\frac{H(x, v)}{H(x', v')}, 1 \right\}
\end{align}
where $H$ is the target Hamiltonian
\begin{align*}
    H(x, v) = -\log p_c(x) + \frac{1}{2} v^T M v.
\end{align*}

We fix the mass matrix $M = I$ and the number of steps $k = 5$. This leaves two parameters, the ste-psize $\epsilon$ and temperature $\lambda$. As with R-MALA we search over $\epsilon \in \{.1, .01, .001\}$ and $\lambda \in \{.5, 1, 2.\}$.

\subsection{Experimental Details}
We compare D-SVGD, R-MALA, and R-HMC to Gibbs-With-Gradients at the task of sampling from RBMs. We present results in two settings; random RBMs with increasing dimension, and an RBM trained on MNIST using Contrastive Divergence. 

The random RBMs have visible dimension $[25, 50, 100, 250, 500, 1000]$ and all have $100$ hidden units. The MNIST RBM has 784 visible units and 500 hidden units and is trained as in Appendix \ref{app:rbm_exp_details}. Following \citet{han2020stein} the random RBMs are initialized as
\begin{align*}
    W \sim N(0, .05I), \qquad b,c \sim N(0, I).
\end{align*}

All samples are initialized to a random uniform Bernoulli distribution and all samplers are run for 2000 steps. We evaluate by computing the Kernelized MMD between each sampler's set of samples and a set of approximate ``ground-truth'' samples generated with the RBMs efficient block-Gibbs sampler. We generate 500 ground truth samples and 100 samples for each sampler tested. In Figure \ref{fig:relax_rbm} we plot the final log-MMD with standard-error over 5 runs with different random seeds. Samples on the right of the figure are generated in the same way from the MNIST RBM.

For D-SVGD we search over relaxation temperatures ${\lambda \in \{.5, 1., 2.\}}$. We optimize the samples with the Adam optimizer~\citep{kingma2014adam}. We search over learning rates in $\{.01, .001, .0001\}$. We use an RBF kernel ${k(x, x') = \exp\left(\frac{−||x−x'||^2}{h}\right)}$ and $h = \text{med}^2/(2 \log(n+ 1))$ where med is the median pairwise distance between the current set of $n$ samples.

All reported results for D-SVGD, R-MALA, and R-HMC are the \emph{best} results obtained over all tested hyper-parameters. We found all of these methods to be very sensitive to their hyper-parameters -- in particular, the relaxation temperature $\lambda$. We believe it may be possible to improve the performance of these methods through further tuning of these parameters but we found doing so beyond the scope of this comparison. 

\section{Gibbs-With-Gradients Extensions}
\label{app:extensions}
\subsection{Extensions To Larger Windows}
We can easily extend our approach to proposals with larger window sizes. This would amount to a a Taylor-series approximation to likelihood ratios where more than 1 dimension of the data has been perturbed. These would come in the form of linear functions of $f(x)$ and $\nabla_x f(x)$. It is likely, of course, that as the window-size is increased, the accuracy of our approximation will decrease as will the quality of the sampler.

In all of our experiments, we found a window-size of 1 to give a considerable improvement over various baselines so we did not explore further. We believe this is an exciting avenue for future work. 

\subsection{Multi-Sample Variant}

As mentioned, all experiments presented in the main paper use a window size of 1 meaning only 1 dimension can be changed at a time. In the binary case, we sample a dimension $i \sim q(i|x)$ which tells us which dimension to flip to generate our proposed update. A simple, and effective extension to this is to simply re-sample multiple indices from this same distribution
$$i_1, \ldots, i_N \sim q(i|x)$$
where $N$ is the number of draws. We then generate $x'$ by flipping the bit at each  sampled index $i_n$. This changes the acceptance probability to
\begin{align}
    \min\left\{ \exp(f(x') - f(x)) \frac{\prod_{n=1}^N q(i_n|x')}{\prod_{n=1}^N q(i_n|x)}, 1\right\}.
\end{align}

\begin{figure}[h]
    \centering
    \includegraphics[height=.25\textwidth]{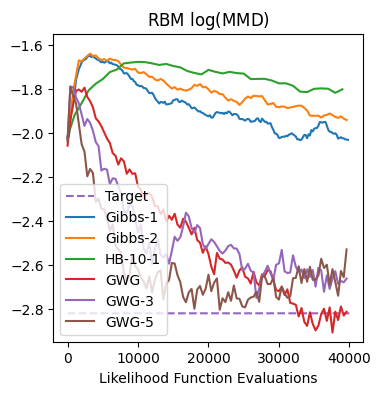}
     \includegraphics[height=.25\textwidth]{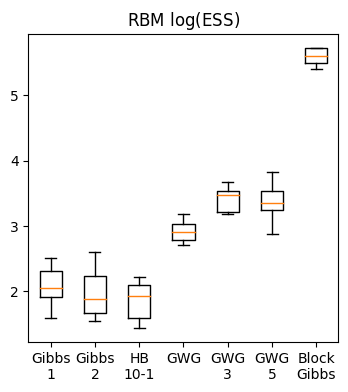}
    \caption{RBM Sampling with Gibbs-With-Gradients extensions. GWG-3 and GWG-5 are the multi-sample variant of GWG described above with $n=3$ and $n=5$, respectively.}
    \label{fig:samp_rbm_3_5}
\end{figure}

This proposal makes a larger number of approximations and assumptions but we find that in some settings it can provide faster convergence and can have reasonable acceptance rates. We demonstrate this in our RBM experiments in Figure \ref{fig:samp_rbm_3_5}. We replicate Figure \ref{fig:samp_rbm} but add the multi-sample variant described above with $N=3$ and $N=5$ samples. We find in this case the multi-sample variant has faster convergence and greater ESS.

\section{Restricted Boltzmann Machines}
\label{app:rbm}

Restricted Boltzmann Machines define a distribution over binary data $x$ and latent variables $h$. The model is defined as:
\begin{align}
     \log p(x, h) = h^T W x + b^Tx + c^Th - \log Z
\end{align}
where $Z$ is the normalizing constant and $\{W, b, c\}$ are the model's parameters. In this model we can efficiently marginalize out the latent variable $h$ to obtain:
\begin{align}
    \log p(x) &= \log \sum_h p(x, h)\nonumber\\
    &= \log \sum_h \exp(h^T W x + b^Tx + c^Th)\nonumber\\
    &= \log (1 + \exp(Wx + c)) + b^Tx - \log Z
\end{align}

While the joint and marginal are both unnormalized, we can see the conditional distirbutions can be easily normalized and take the form:
\begin{align}
    p(x|h) &= \text{Bernoulli}(Wx + c)\nonumber\\
    p(h|x) &= \text{Bernoulli}(W^Th + b)\nonumber.
\end{align}

We can exploit this structure to more efficiently sample from RBMs. We can perform Block-Gibbs updates by starting at some initial $x$, and repeatedly sample $h\sim p(h|x)$, ${x \sim p(x|h)}$. Exploiting this structure leads to much more efficient sampling than standard Gibbs and other samplers (see Figure \ref{fig:samp_rbm}).

\subsection{Experimental Details}
\label{app:rbm_exp_details}
We explore the performance of various approaches to sample from an RBM trained on the MNIST dataset. The RBM has 500 hidden units (and 784 visible units). We train the model using contrastive divergence~\citep{hinton2002training} for 1 epoch through the dataset using a batch size of 100. We use 10 steps of MCMC sampling using the Block-Gibbs sampler defined above to generate samples for each training iteration. We use the Adam~\citep{kingma2014adam} optimizer with a learning rate of $.001$. 

Our first result compares samples generated by various approaches with samples generated with the Block-Gibbs sampler described above. We generate a set of 500 samples using the Block-Gibbs sampler run for 10,000 steps. At this length, the samples are very close to true samples from the model. Next we generate a set of 100 samples from a number of other samplers: Gibbs, Hamming Ball and Gibbs-With-Gradients. After every MCMC transition we compute the Kernelized Maximum Mean Discrepancy~\citep{gretton2012kernel} between the current set of samples and our ``ground-truth'' long-run Block-Gibbs samples. We use an exponential average Hamming kernel $K(x, x') = \exp\left(-\sum_{i=1}^D \frac{\mathbf{1}(x_i = x'_i)}{D}\right)$ to compute the MMD.

The next result is the effective sample size of a test statistic for each sampler. Following \citet{zanella2020informed}, our test statistic is the Hamming distance between the current sample and a random input configuration. We present a box-plot showing the median, standard-deviation, and outliers over 32 chains.

\subsection{Additional Results}
\begin{figure}[h]
    \centering
    \includegraphics[width=.45\textwidth]{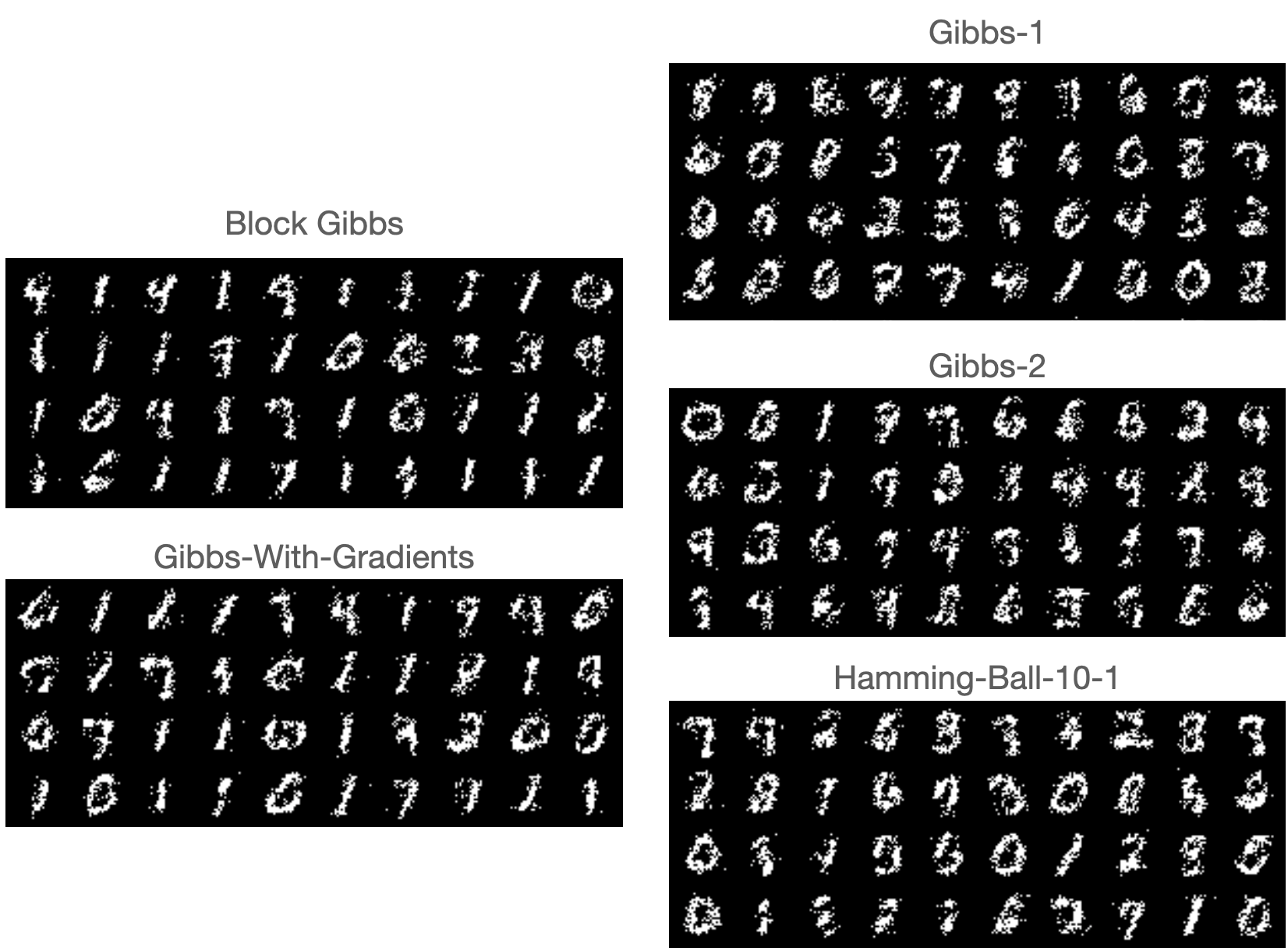}
    \caption{Sets of samples drawn from a fixed RBM with various MCMC approaches after 10,000 steps.}
    \label{fig:rbm_samples}
\end{figure}

We visualize the samples after 10,000 steps of each tested sampler in Figure \ref{fig:rbm_samples}. We can see the Gibbs-With-Gradients samples much more closely matches the Block-Gibbs samples. This result is reflected in the improved MMD scores see in Figure \ref{fig:samp_rbm} (left).

\section{Ising Models}
\label{app:ising}
Ising models are unnormalized models for binary data defined as
\begin{align}
    \log p(x) = x^T J x + b^Tx - \log Z
\end{align}
where $J$ and $b$ are the model's parameters and $Z$ is the normalizing constant. $J$ determines which other variables each $x_i$ is correlated with. If $J=0$ then the model becomes a factorized Bernoulli distribution. If all of the non-zero indices of $J$ are the same, then we can pull out this value and rewrite the model as
\begin{align}
    \log p(x) = \theta x^T J x + b^Tx - \log Z
\end{align}
where now $\theta$ controls how correlated each $x_i$ is with its connected variables and $J$ controls which variables each $x_i$ is connected to. Our lattice Ising models take this form where the $J$ is the adjacency matrix of a cyclic 2D lattice and $\theta$ controls the strength of the connectivity. 

\subsection{Experimental Details: Sampling}
We experiment with our sampler's ability to sample from Ising models on the 2D cyclic lattice as various factors chage. These include the connectivity strength and the size of the lattice. We run each sampler for 100,000 steps and evaluate using the ESS of a test statistic. Following \citet{zanella2020informed} our test statistic is the Hamming distance between the current sample and a random input configuration. We present the ESS (in log-scale), averaged with standard-errors, over 32 random seeds. 

We can see in both 10x10 and 40x40 lattice sizes, our sampler outperforms Gibbs and the Hamming ball.

\subsection{Experimental Details: Training}
\label{app:tr_ip}
We create Ising models with 2 kinds of graph structure; a cyclic 2D lattice and a random Erdos-Renyi (ER) graph. For the lattice we create models with a 10x10, 25x25, and 40x40 lattice leading to 100, 625, and 1600 dimensional distributions. We train models with connectivity $\theta \in [-.1, 0.0, .1, .25, .5]$. 

For the ER graph, we create a model with 200 nodes. The ER edge probability is chosen so each node has an average of 4 neighbors. The strength of each edge is IID sampled from $N\left(0, \frac{1}{4}\right)$.

A dataset of 2000 examples is generated from each model using 1,000,000 steps of Gibbs sampling. We train models using persistent contrastive divergence~\citep{tieleman2008training} with a buffer size of 256 examples. Models are trained with the Adam optimizer~\citep{kingma2014adam} using a learning rate of .001 and a batch size of 256. We update the persistent samples using Gibbs and Gibbs-With-Gradients. We train models with $\{5, 10, 25, 50, 100\}$ steps of MCMC per training iteration and compare their results. We train all models with an $\ell1$ penalty to encourage sparsity with strength .01. 

We compare results using the root-mean-squared-error between the true connectivity matrix $J$ and the inferred connectivity matrix $\hat{J}$.

\subsection{Additional Results: Training Ising Models}
\label{app:tr_ip_add}
\begin{figure}[h]
    \centering
    \includegraphics[height=.22\textwidth]{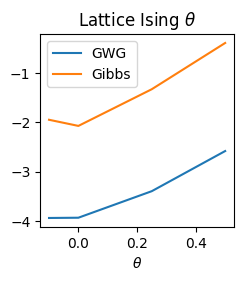}
     \includegraphics[height=.22\textwidth]{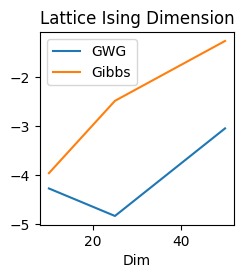}
      \includegraphics[height=.22\textwidth]{figs/LOG_STEPS_rmse_sigma_0.25_dim_25_LABEL.png}
     \includegraphics[height=.22\textwidth]{figs/LOG_rmse_er_ising.png}
    \caption{Training Ising models. Top Left: Lattice Ising with increasing $\theta$ (dim = 50, steps = 50). Top Right: Lattice Ising with increasing dimension ($
    \theta=.25$, steps = 25). Bottom Right: Lattice Ising with increasing steps (dim = 25, $\theta = .25$). Bottom Right: Erdos-Renyi Ising with increasing steps. Values are $\log$(RMSE) between the learned $J$ and the true $J$. Gibbs-With-Gradients leads to better solutions with lower computational cost. }
    \label{fig:tr_ip_add}
\end{figure}

In Figure \ref{fig:tr_ip_add}, we present an expanded version of Figure \ref{fig:tr_ip} which presents additional results. In these additional experiments we find Gibbs-With-Gradients considerably outperforms training with Gibbs sampling. 

\section{Factorial Hidden Markov Models}
\label{app:fhmm}
Factorial Hidden Markov Models (FHMM) are a generalization of Hidden Markov Models and model real-valued time-series data. The observed data $y \in R^L$ is generated conditioned on a discrete latent-variable $x \in \{0,1\}^{L\times K}$. This latent-variable is drawn from the product of $K$ independent Markov processes as seen below. The data $y_t$ is generated by by the $K$-dimensional state vector $x_t$ with Gaussian noise added. 
\begin{align}
    p(x, y) &= p(y|x)p(x)\nonumber\\
    p(y|x) &=   \prod_{t=1}^L  \mathcal{N}(y_t; Wx_t + b, \sigma^2)\nonumber\\
    p(x) &= p(x_1)\prod_{t=2}^L p(x_t|x_{t-1})\nonumber\\
    p(x_1) &= \prod_{k=1}^K \text{Bernoulli}(x_{1k}; \alpha_k)\nonumber\\
    p(x_{t+1}|x_t) &= \prod_{k=1}^K \text{Bernoulli}(x_{(t+1)k}; \beta_k^{x_{tk}}(1-\beta_k)^{1-x_{tk}})
\end{align}
The posterior $p(x|y)$ has no closed form and thus we must rely on MCMC techniques to sample from it.

\subsection{Experimental Details}
We sample the parameters of an FHMM randomly as
\begin{align}
    W, b &\sim \mathcal{N}(0, I)
\end{align}
and set $\sigma^2 = .5$, $\alpha_k = .1$ and $\beta_k = .95$ for for all $k$.

We then sample $x \sim p(x)$ and $y \sim p(y|x)$ and run all samplers for 10,000 steps to generate samples from $p(x|y)$. The Hamming Ball Sampler~\citep{titsias2017hamming} is special for this model as it exploits the known block-structure of the posterior. We use a block size of 10 and the blocks are chosen to be all 10 dims of the latent state at a randomly chosen time $x_t$. Thus, this sampler is aware of more hard-coded structure in the model than the Gibbs baseline and Gibbs-With-Gradients.

\section{Potts Models for Proteins}
\label{app:tr_prot}
We train the MCMC models using PCD~\citep{tieleman2008training} with a buffer size of 2560. At each training iteration we sample a mini batch of 256 examples and 256 random samples from the persistent sample buffer. These are updated using 50 steps of either Gibbs or Gibbs-With-Gradients and the gradient estimator of Equation \ref{eq:ml_grad_est} is used to update the model parameters. We train for 10,000 iterations using the Adam optimizer~\citep{kingma2014adam}. Following \citet{marks2011protein} we use block-$\ell1$ regularization. This regularizer takes the form
\begin{align}
    \mathcal{L}_1(J) = \sum_{ij} ||J_{ij}||_2.
\end{align}

We add this regularizer to the maximum likelihood gradient estimator. We tested regularization strength parameters in $\{.1, .03, .01\}$ and found $.01$ to work best for PLM, Gibbs, and Gibbs-With-Gradients.

Ground truth couplings were extracted from an experimentally validated distance-map. As is standard, we consider any pair of amino acids to be a contact if they are within 5 angstroms of each other. 

\subsection{Recall on PF10018}
\begin{figure}[h]
    \centering
    \includegraphics[width=.23\textwidth]{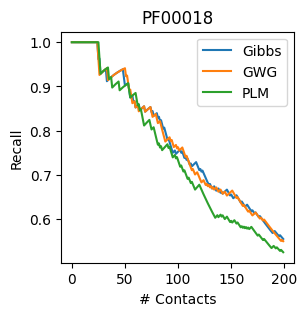}
    \caption{Recall Curves for contact prediction with Potts models.}
    \label{fig:tr_prot_pf}
\end{figure}

We do not present results on PF10018 in the main body as it was used to tune hyper-parameters. For completeness, we present them here in Figure \ref{fig:tr_prot_pf}. As with the other protiens, the MCMC-based training outperforms PLM but by a smaller margin and GWG and Gibbs perform comparably here. This further supports the benefit of MCMC training over PLM sets in on larger data as does the benefit of GWG over Gibbs.

\subsection{Visualized Contacts}
\begin{figure}[h]
    \centering
    \includegraphics[width=.23\textwidth]{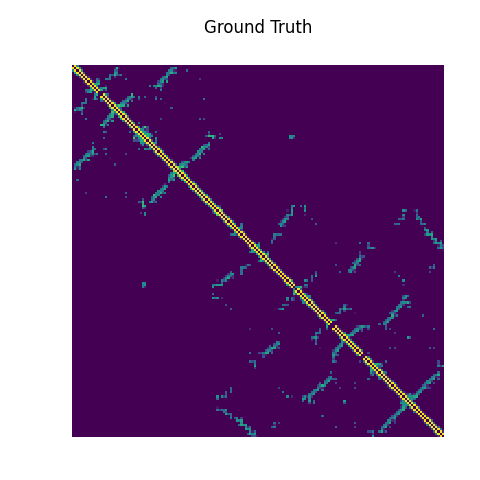}
    \includegraphics[width=.23\textwidth]{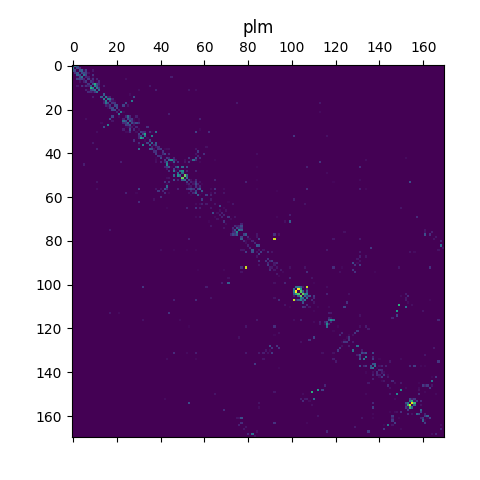}
    \includegraphics[width=.23\textwidth]{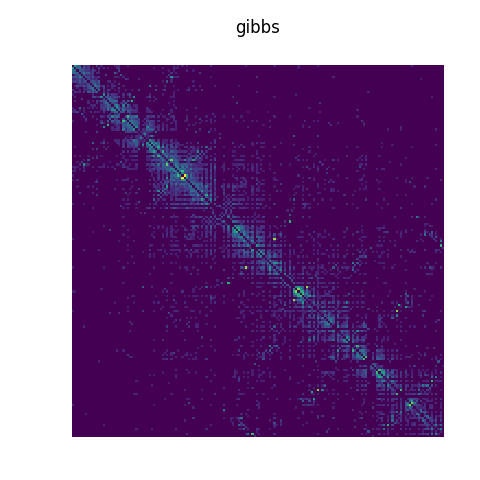}
    \includegraphics[width=.23\textwidth]{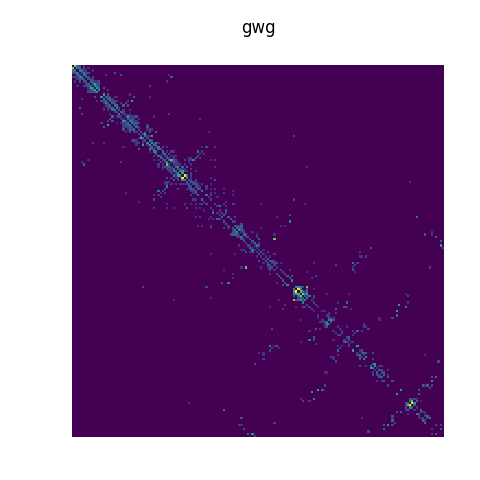}
    
    \caption{Inferred Couplings for CADH1\_HUMAN. ``Ground Truth'' is the matrix of known distances between amino acids. All other matrices are the norms of the Potts model $J_{ij}$ parameter.}
    \label{fig:tr_coup}
\end{figure}

We visualize the inferred couplings for CADH1\_HUMAN in Figure \ref{fig:tr_coup}. We see that GWG most accurately matches the known structure with Gibbs inferring spurious couplings and PLM missing many couplings near the diagonal.

\section{Deep EBMs}
\label{app:deep_ebm}

\subsection{Architectures}

We train on two types of data; binary and categorical. For the binary datasets, the data is represented as $\{0, 1\}^D$ where $D$ is the dimension of the data. 

For the categorical datasets, each categorical variable is represented as a ``one-hot'' vector. Thus, for image data, each pixel is represented with a 256-dimensional vector. To deal with the very high dimensionality of this input parameterization, we first map each one-hot vector to a learned, low-dimensional embedding with a linear transformation. We map to $D_p = 4$ dimensions for all models tested. We then feed this $(D \times D_p)$-dimensional input to our network. There are certainly more efficient ways to represent this data, but our intention was not to achieve state-of-the-art results on these datasets and instead to demonstrate our sampler could enable the training of energy-based models on high-dimensional discrete data, so we use the most straight-forward parameterization.

The ResNet used for our EBM is identical for all datasets. The network has 8 residual blocks with 64 feature maps. Each residual block has 2 convolutional layers. The first two residual blocks have a stride of 2. The output features are mean-pooled across the spatial dimensions and a single linear layer is used on top to provide the energy. The Swish~\cite{ramachandran2017searching} nonlinearity ($x\cdot \sigma(x)$) is used throughout.

\subsection{Experimental Details}
We trained all models Adam~\cite{kingma2014adam} using a learning rate of $0.0001$. We linearly warm-up the learning rate for the first 10,000 iterations. We found this was necessary to help burn in the replay buffer of samples. 

For the large datasets (static/dynamic MNIST, Omniglot) we use a replay buffer with 10,000 samples. For the smaller datasets (Caltech, Freyfaces, Histopathology) the buffer size is 1000. Unlike recent work on continuous EBMs~\citep{du2019implicit, grathwohl2019your}, we do not reinitialize the buffer samples to noise. We found this resulted in unstable training and lower likelihoods.

We train all models for 50,000 iterations. We use the same training/validation/testing splits as \citet{tomczak2018vae}. We evaluate models every 5000 iterations using 10,000 steps of AIS. We select the model which performs the best on the validation data under this procedure. Final results in Table \ref{tab:ebm_ll} are generated from the selected models by running 300,000 iterations of AIS. We evaluate using a model who's weights are given by an exponential moving average of the training model's weights. This is analogous to training with ``fast-weights'' as in \citet{tieleman2009using}. We find this greatly improves likelihood performance and sample quality. We use an exponential moving average with weight $0.999$ and did not experiment with other values. 

We believe better results could be obtained with larger models or alternative architectures, but we leave this for future work.  

\subsubsection{Partition Function Estimation with AIS}
\label{app:part_est}

We estimate likelihoods by estimating the partition function using Annealed Importance Sampling (AIS)~\citep{neal2001annealed}. AIS underestimates the log-partition-function leading to \emph{over-estimating} the likelihood. The estimation error can be reduced by using a larger number of intermediate distributions or a more efficient MCMC sampler. Results presented in Table \ref{tab:ebm_ll} were generated with 300,000 intermediate distributions. We chose this number as it appears to be sufficiently high for our partition function estimate to converge. Despite this, these are upper-bounds and therefore should not be considered definitive proof that one model achieves higher likelihoods than another. 

We anneal between our model's unnormalized log-probability $f(x)$ and a multivariate Bernoulli or Categorical distribution, $\log p_n(x)$, fit to the training data, for binary and categorical data, respectively. 

\begin{align}
f_t(x) = \beta_t f(x) + (1 - \beta_t)\log p_n(x)
\end{align}
where $\beta_t$ is linearly annealed from 0 to 1. Alternative strategies such as sigmoid annealing could be used, but we leave this for future work.

\begin{figure}[h]
    \centering
    \includegraphics[width=.28\textwidth]{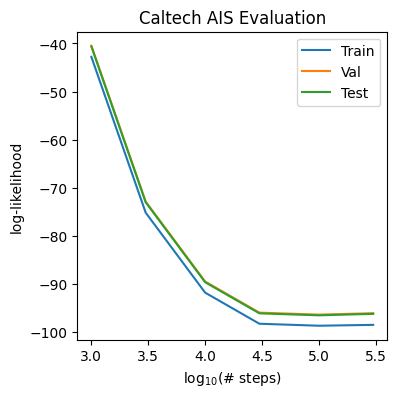}
    \caption{AIS likelihood estimates as the number of intermediate distributions increases for our Caltech Silhouettes Resnet EBM. Values converge after $30,000 \approx 10^{4.5}$ step s}
    \label{fig:ais}
\end{figure}

In Figure \ref{fig:ais} we plot the estimated likleihoods for our Caltech Silhouettes models as the number of intermediate distributions increases. It can be seen that between 30,000 and 300,000 ($\approx 10^{4.5} \rightarrow 10^{5.5}$) the values appear to be converged, thus we feel our reported number faithfully represent our models' performance.

\subsection{Additional Results}
We present additional long-run samples from our convolutional EBM. These samples were generated using an annealed Markov Chain (as described above) and Gibbs-With-Gradients as the base MCMC sampler. 

\begin{figure}[h]
    \centering
    \includegraphics[width=.45\textwidth]{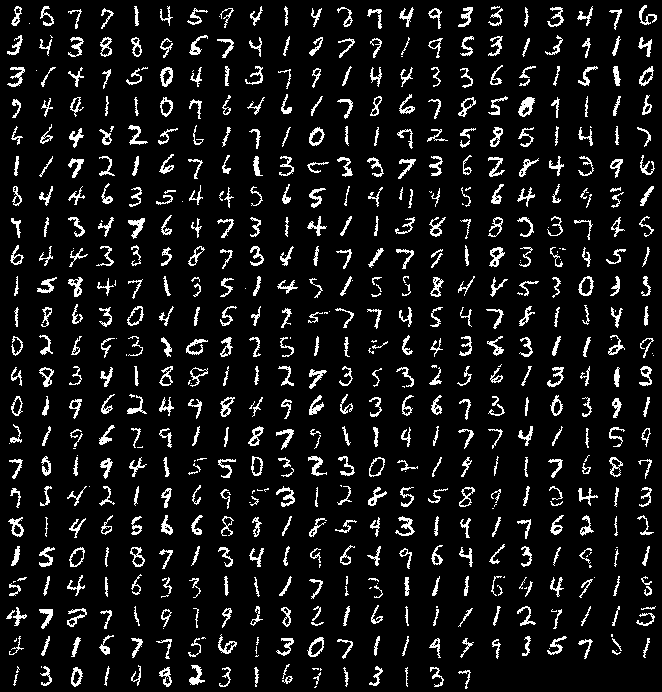}
    \caption{Static MNIST Samples}
    \label{fig:smnist_samples}
\end{figure}

\begin{figure}[h]
    \centering
    \includegraphics[width=.45\textwidth]{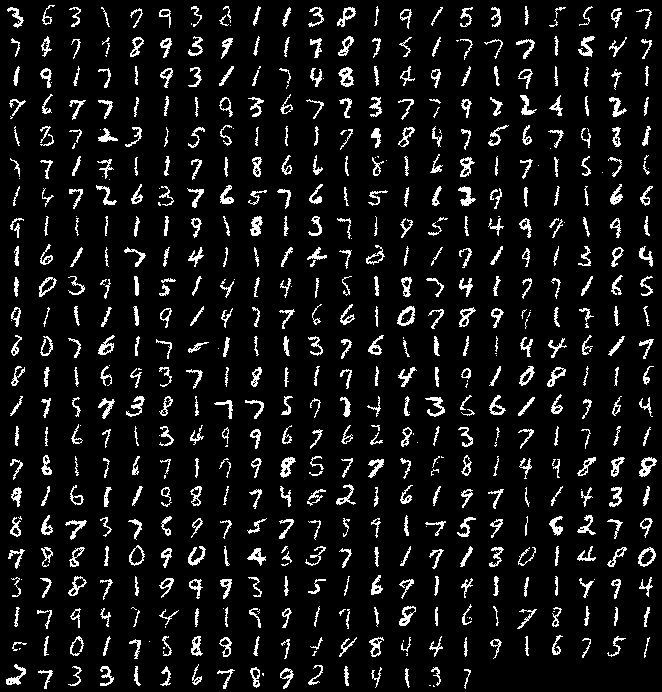}
    \caption{Dynamic MNIST Samples}
    \label{fig:dmnist_samples}
\end{figure}

\begin{figure}[h]
    \centering
    \includegraphics[width=.45\textwidth]{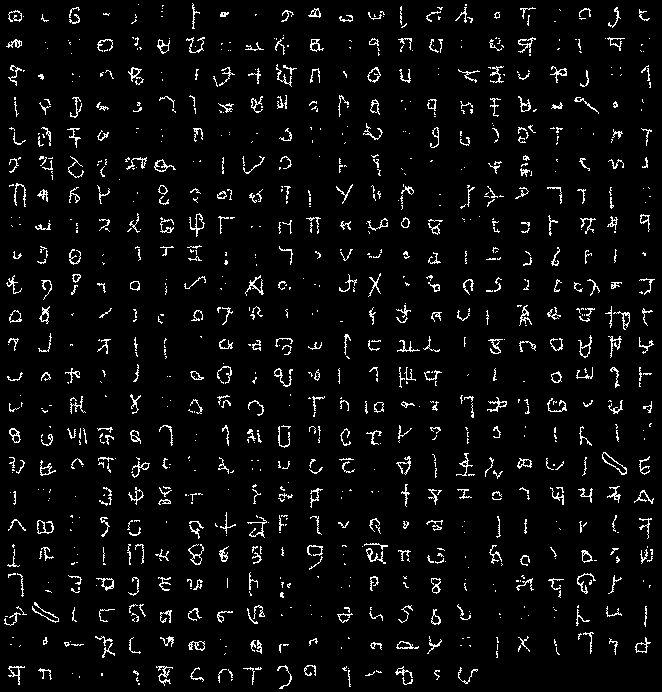}
    \caption{Omniglot Samples}
    \label{fig:omni_samples}
\end{figure}

\begin{figure}[h]
    \centering
    \includegraphics[width=.45\textwidth]{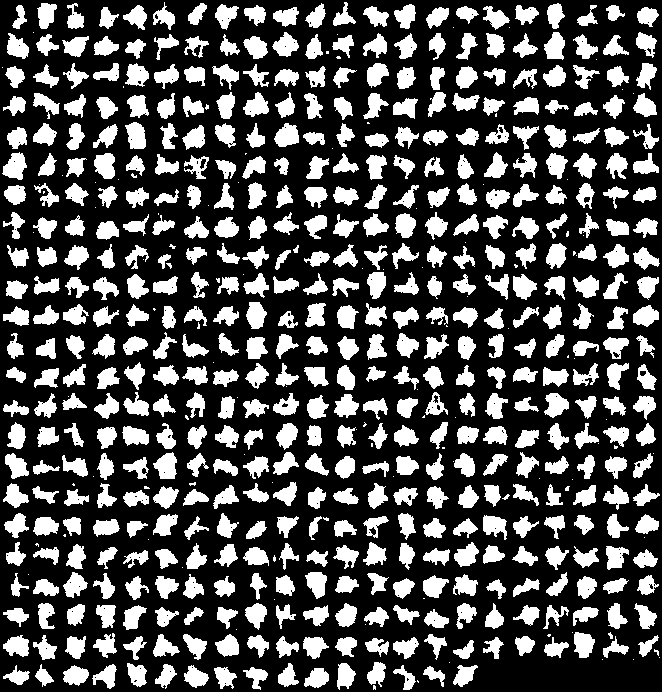}
    \caption{Caltech Silhouette Samples}
    \label{fig:cal_samples}
\end{figure}

\begin{figure}[h]
    \centering
    \includegraphics[width=.4\textwidth]{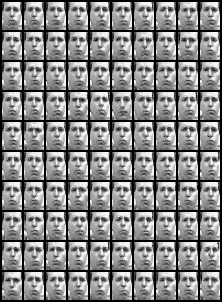}
    \caption{Freyfaces Samples}
    \label{fig:cal_samples}
\end{figure}

\begin{figure}[h]
    \centering
    \includegraphics[width=.45\textwidth]{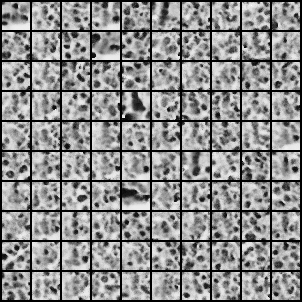}
    \caption{Histopathology Samples}
    \label{fig:cal_samples}
\end{figure}
\clearpage

\section{Preliminary Text EBM Results}
\label{app:text}
There has recently been interest in learning Energy-Based Models of text data. An EBM for text could enable non-autoregressive generation and more flexible conditioning than autoregressive models. For example, an EBM trained on language pairs $p(x, y)$ could be used to translate in either direction without retraining and could be structured as $\log p(x, y) = f_\theta(x)^T g_\phi(y) - \log Z$ so that each language component could be trained separately. We also have much more architectural freedom when specifying EBMs meaning $f_\theta$ is free to be a CNN, RNN, transformer, or MLP.

A few works have had success training and applying text EBMs. \citet{deng2020residual} find that EBMs can be used to improve the generative modeling performance of large-scale transformer language models and \citet{he2021joint} find that Joint Energy-Based Models~\citep{grathwohl2019your} can improve the calibration of text classifiers. Because of the discrete structure of the text data, both of these works train using Noise Contrastive Estimation~\citep{gutmann2010noise} using a pretrained autoregressive language model as the noise distribution. NCE requires a noise distribution which can be sampled from and enables exact likelihood computation. Thus, these approaches rely on and are limited by the quality of these autoregressive models.

Ideally, we could train a text EBM on its own, without an auxiliary model. One way to do this is to use the gradient estimator of Equation \ref{eq:ml_grad_est} but the MCMC sampling task is very challenging. Text models typically have a vocabulary above 10,000 words so the size of the discrete sample space is tremendous. Further, as noted in Section \ref{sec:deep}, to apply Gibbs sampling to a model like this we would need to evaluate the energy function over 10,000 times to perform a single step of sampling!

We believe Gibbs-With-Gradients can provide an avenue to train and sample from these kinds of models. As a preliminary experiment we train non-autoregressvive language models on a shortened version of the Penn Tree Bank dataset~\citep{taylor2003penn}. This is a dataset of short sentences with 10,000 words. We cut out all sentences with greater than 20 words and pad all shorter sentences with an ``end of sequence'' token. We feel this simplified setting is sufficient for a proof-of-concept as the configuration space is very large and Gibbs sampling is not feasible.

Our model consists of a bidirectional LSTM~\citep{gers1999learning} with 512 hidden units. We project the 10,000 words to an embedding of size 256 with a learned mapping. To compute the energy, we take the last hidden-state from each direction, concatenate them together to a 1024-dimensional vector. We project this to 512 dimensions with a linear layer, apply a Swish nonlinearity and then map to 1 dimension with another linear layer. 

We train using PCD with a buffer size of 1000 and we use 20 steps of MCMC with Gibbs-With-Gradients to update the samples at every training iteration. Besides this, training was identical to our image EBMs in section \ref{sec:deep}.

We compare with a simple autoregressive language model which is based on an LSTM with 512 hidden units and use a learned word embedding of size 256. 

We find the autoregressive model slightly outperforms the EBM. The test-set log-likelihoods of the EBM and autoregressive model are $-77.16$ and $-74.0$, respectively. For comparison, a uniform distribution over possible tokens obtains $-184.21$ and a Categorical distribution fit to the training data obtains $-100.05$.




While we are aware these are \emph{far} from state-of-the-art language modelling results, we believe they demonstrate that Gibbs-With-Gradients can enable MCMC-trained EBMs to successfully model text data with large vocabulary sizes. At every step, the sampler has $10,000\times 20 = 200,000$ choices for possible updates. Despite this massive sampling space, we find our acceptance rates during training are just above 70\% making our approach at least 3500 times more efficient than Gibbs sampling.

We believe improvements could be obtained through larger models and more tuning. To further scale this approach, we believe we will need to develop further approximations which make sampling from very large categorical distributions more efficient and numerically stable. We leave this for future work.

\end{document}